\documentclass[10pt,twocolumn,letterpaper]{article}

\usepackage{cvpr}
\usepackage{times}
\usepackage{epsfig}
\usepackage{graphicx}
\usepackage{amsmath}
\usepackage{amssymb}
\usepackage{appendix}
\usepackage{xcolor}
\usepackage{overpic}
\usepackage{tabulary,multirow}
\usepackage[caption=false]{subfig}
\usepackage{makecell}
\usepackage{setspace}
\PassOptionsToPackage{hyphens}{url}\usepackage[pagebackref=true,breaklinks=true,letterpaper=true,colorlinks,
  citecolor=citecolor,bookmarks=false]{hyperref}
\definecolor{citecolor}{RGB}{34,139,34}
\def\cvprPaperID{}\cvprfinalcopy
\newcolumntype{x}[1]{>{\centering\arraybackslash}p{#1pt}}
\newlength\savewidth\newcommand\shline{\noalign{\global\savewidth\arrayrulewidth
  \global\arrayrulewidth 1pt}\hline\noalign{\global\arrayrulewidth\savewidth}}
\newcommand{\tablestyle}[2]{\setlength{\tabcolsep}{#1}\renewcommand{\arraystretch}{#2}\centering\footnotesize}
\makeatletter\renewcommand\paragraph{\@startsection{paragraph}{4}{\z@}
  {.3em \@plus1ex \@minus.2ex}{-.4em}{\normalfont\normalsize\bfseries}}\makeatother
\newcommand{\posc}{\mathcal{P}_c}
\newcommand{\negc}{\mathcal{N}_c}
\newcommand{\vocab}{\mathcal{V}}
\newcommand{\allimgs}{\mathcal{D}}
\newcommand{\img}{i}
\newcommand{\exhaustive}{e_{\img}^c}
\newcommand{\lvis}{\textsc{LVIS}\xspace}
\newcommand{\app}{\raise.17ex\hbox{$\scriptstyle\sim$}}
\newcommand{\h}{\kern-.1ex\raise.17ex\hbox{--}\kern.5ex}
\newcommand{\eve}{expert\kern.2ex1 \vs expert\kern.2ex2}
\definecolor{demphcolor}{RGB}{100,100,100}
\newcommand{\demph}[1]{\textcolor{demphcolor}{#1}}
\newcommand{\train}{\texttt{train}\xspace}
\newcommand{\val}{\texttt{val}\xspace}
\newcommand{\test}{\texttt{test}\xspace}
\newcommand{\mypm}[1]{{\tiny{{\demph{{$\pm$#1}}}}}}

\makeatletter
\newcommand\footnoteref[1]{\protected@xdef\@thefnmark{\ref{#1}}\@footnotemark}
\makeatother

\begin{document}

%%%%%%%%%%%%%%%%%%%%%%%%%%%%%%%%%%%%%%%%%%%%%%%%%%%%%%%%%%%%%%%%%%%%%%%%%%%%%%%%%%%%%%%%%%%%%%%%%%%
\title{\lvis: A Dataset for Large Vocabulary Instance Segmentation\vspace{-2mm}}
\author{%
 Agrim Gupta \quad Piotr Doll\'ar \quad Ross Girshick \vspace{.5em}\\
 Facebook AI Research (FAIR) \vspace{.5em}
}
\maketitle
%\thispagestyle{empty} % uncomment to remove page 1 number

%%%%%%%%%%%%%%%%%%%%%%%%%%%%%%%%%%%%%%%%%%%%%%%%%%%%%%%%%%%%%%%%%%%%%%%%%%%%%%%%%%%%%%%%%%%%%%%%%%%
\begin{abstract}
\vspace{-2.5mm}
Progress on object detection is enabled by datasets that focus the research community's attention on open challenges. This process led us from simple images to complex scenes and from bounding boxes to segmentation masks. In this work, we introduce \lvis (pronounced `el-vis'): a new dataset for Large Vocabulary Instance Segmentation. We plan to collect \app 2 million high-quality instance segmentation masks for over 1000 entry-level object categories in 164k images. Due to the Zipfian distribution of categories in natural images, \lvis naturally has a long tail of categories with few training samples. Given that state-of-the-art deep learning methods for object detection perform poorly in the low-sample regime, we believe that our dataset poses an important and exciting new scientific challenge. \lvis is available at \url{http://www.lvisdataset.org}.
\end{abstract}
\vspace{-2mm}

%%%%%%%%%%%%%%%%%%%%%%%%%%%%%%%%%%%%%%%%%%%%%%%%%%%%%%%%%%%%%%%%%%%%%%%%%%%%%%%%%%%%%%%%%%%%%%%%%%%
\section{Introduction}\label{sec:intro}
A central goal of computer vision is to endow algorithms with the ability to intelligently describe images. Object detection is a canonical image description task; it is intuitively appealing, useful in applications, and straightforward to benchmark in existing settings. The accuracy of object detectors has improved dramatically and new capabilities, such as predicting segmentation masks and 3D representations, have been developed. There are now exciting opportunities to push these methods towards new goals.

Today, rigorous evaluation of general purpose object detectors is mostly performed in the few category regime (\eg 80) or when there are a large number of training examples per category (\eg 100 to 1000+). Thus, there is an opportunity to enable research in the natural setting where there are a large number of categories \emph{and} per-category data is sometimes scarce. \emph{The long tail of rare categories is inescapable; annotating more images simply uncovers previously unseen, rare categories (see Fig.~\ref{fig:analysis:cumulative_category_count} and~\cite{zipf2013psycho,spain2007measuring,russell2008labelme,xiao2010sun})}. Efficiently learning from few examples is a significant open problem in machine learning and computer vision, making this opportunity one of the most exciting from a scientific and practical perspective. But to open this area to empirical study, a suitable, high-quality dataset and benchmark is required.

%##################################################################################################
\begin{figure}[t]
\begin{center}
\includegraphics[height=2.5in,width=\columnwidth]{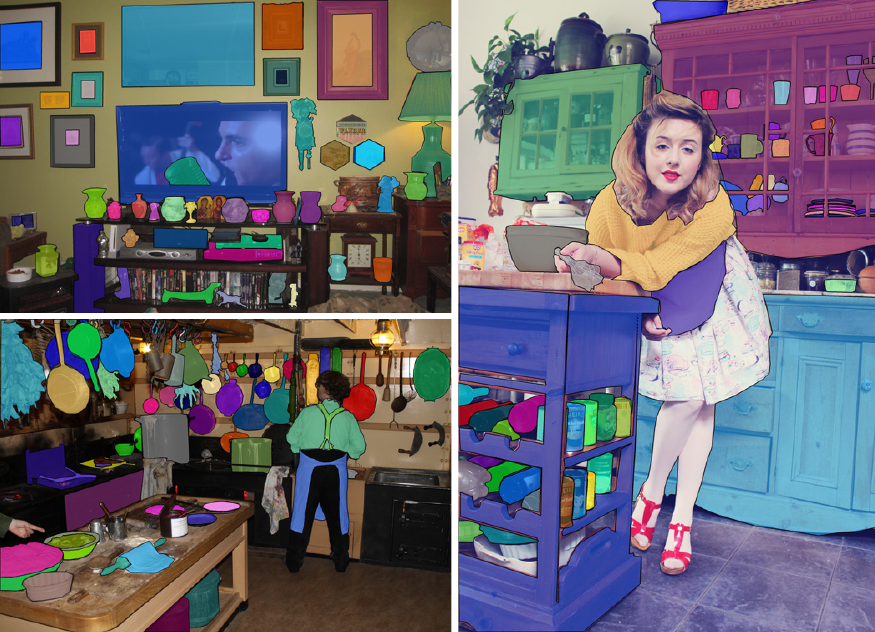}
\end{center}
\vspace{-2mm}
\caption{\textbf{Example annotations.} We present \textbf{\lvis}, a new dataset for benchmarking Large Vocabulary Instance Segmentation in the 1000+ category regime with a challenging long tail of rare objects.}
\label{fig:pull}
\vspace{-4mm}
\end{figure}
%##################################################################################################

We aim to enable this new research direction by designing and collecting \textbf{\lvis} (pronounced `el-vis')---a benchmark dataset for research on Large Vocabulary Instance Segmentation. We are collecting instance segmentation masks for more than 1000 entry-level object categories (see Fig.~\ref{fig:pull}). When completed, we plan for our dataset to contain 164k images and \app 2 million \emph{high-quality} instance masks.\footnote{We plan to annotate the 164k images in COCO 2017 (we have permission to label \texttt{test2017}); \app 2M is a projection after labeling 85k images.} Our annotation pipeline starts from a set of images that were collected without prior knowledge of the categories that will be labeled in them. We engage annotators in an iterative object spotting process that uncovers the long tail of categories that naturally appears in the images and avoids using machine learning algorithms to automate data labeling.

We designed a crowdsourced annotation pipeline that enables the collection of our large-scale dataset while also yielding high-quality segmentation masks. Quality is important for future research because relatively coarse masks, such as those in the COCO dataset~\cite{Lin2014}, limit the ability to differentiate algorithm-predicted mask quality beyond a certain, coarse point. When compared to expert annotators, our segmentation masks have higher overlap and boundary consistency than both COCO and ADE20K~\cite{zhou2016semantic}.

To build our dataset, we adopt an \emph{evaluation-first design principle}. This principle states that we should first determine exactly how to perform quantitative evaluation and only then design and build a dataset collection pipeline to gather the data entailed by the evaluation. We select our benchmark task to be COCO-style instance segmentation and we use the same COCO-style average precision (AP) metric that averages over categories and different mask intersection over union (IoU) thresholds~\cite{cocoap}. Task and metric continuity with COCO reduces barriers to entry.

\begin{figure}[t]
\begin{center}
\includegraphics[width=1.0\columnwidth]{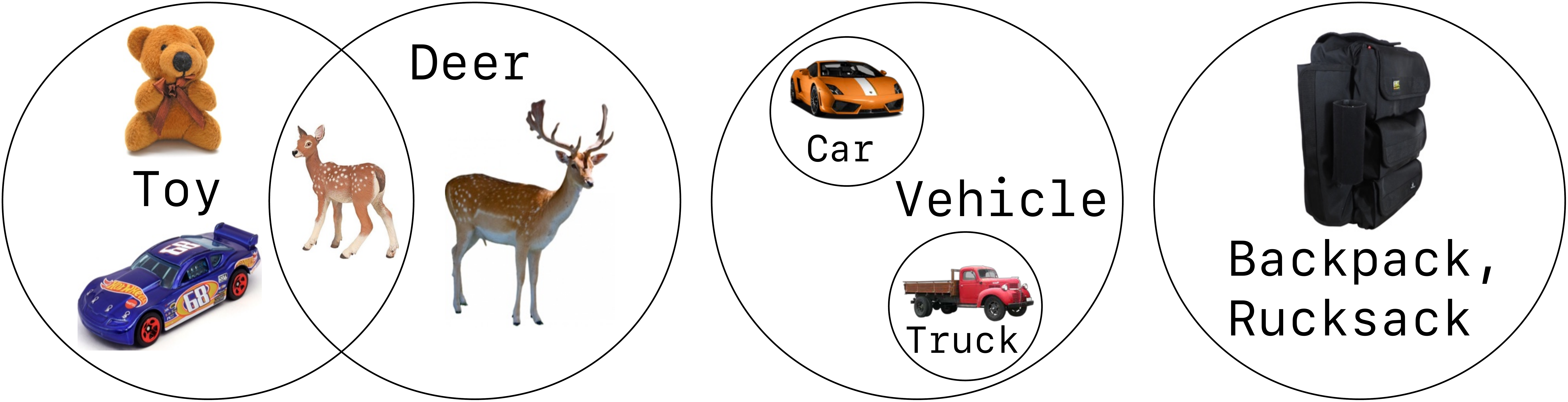}
\end{center}
\vspace{-2mm}
\caption{\textbf{Category relationships from left to right:} non-disjoint category pairs may be in \emph{partially overlapping}, \emph{parent-child}, or \emph{equivalent (synonym)} relationships, implying that a single object may have multiple valid labels. The fair evaluation of an object detector must take the issue of multiple valid labels into account.}
\label{fig:category_venn}
\vspace{-2mm}
\end{figure}

Buried within this seemingly innocuous task choice are immediate technical challenges: How do we fairly evaluate detectors when one object can reasonably be labeled with multiple categories (see Fig.~\ref{fig:category_venn})? How do we make the annotation workload feasible when labeling 164k images with segmented objects from over 1000 categories?

The essential design choice resolving these challenges is to build a \emph{federated dataset}: a single dataset that is formed by the union of a large number of smaller constituent datasets, each of which looks exactly like a traditional object detection dataset for a single category. Each small dataset provides the essential guarantee of \emph{exhaustive annotations} for a single category---\emph{all instances of that category are annotated}. Multiple constituent datasets may overlap and thus a single object within an image can be labeled with multiple categories. Furthermore, since the exhaustive annotation guarantee only holds within each small dataset, we do not require the entire federated dataset to be exhaustively annotated with all categories, which dramatically reduces the annotation workload. Crucially, at test time the membership of each image with respect to the constituent datasets is not known by the algorithm and thus it must make predictions as if all categories will be evaluated. The evaluation oracle evaluates each category fairly on its constituent dataset.

In the remainder of this paper, we summarize how our dataset and benchmark relate to prior work, provide details on the evaluation protocol, describe how we collected data, and then discuss results of the analysis of this data.

\paragraph{Dataset Timeline.} We report detailed analysis on the 5000 image \val subset that we have annotated twice. We have now annotated an additional 77k images (split between \train, \val, and \test), representing \app 50\% of the final dataset; we refer to this as \textbf{\lvis v0.5} (see \S\ref{v0.5} for details). The first \lvis Challenge, based on v0.5, will be held at the COCO Workshop at ICCV 2019.

%##################################################################################################
%##################################################################################################

%##################################################################################################
\begin{figure*}[t]%
\centering
\includegraphics[width=0.99\textwidth]{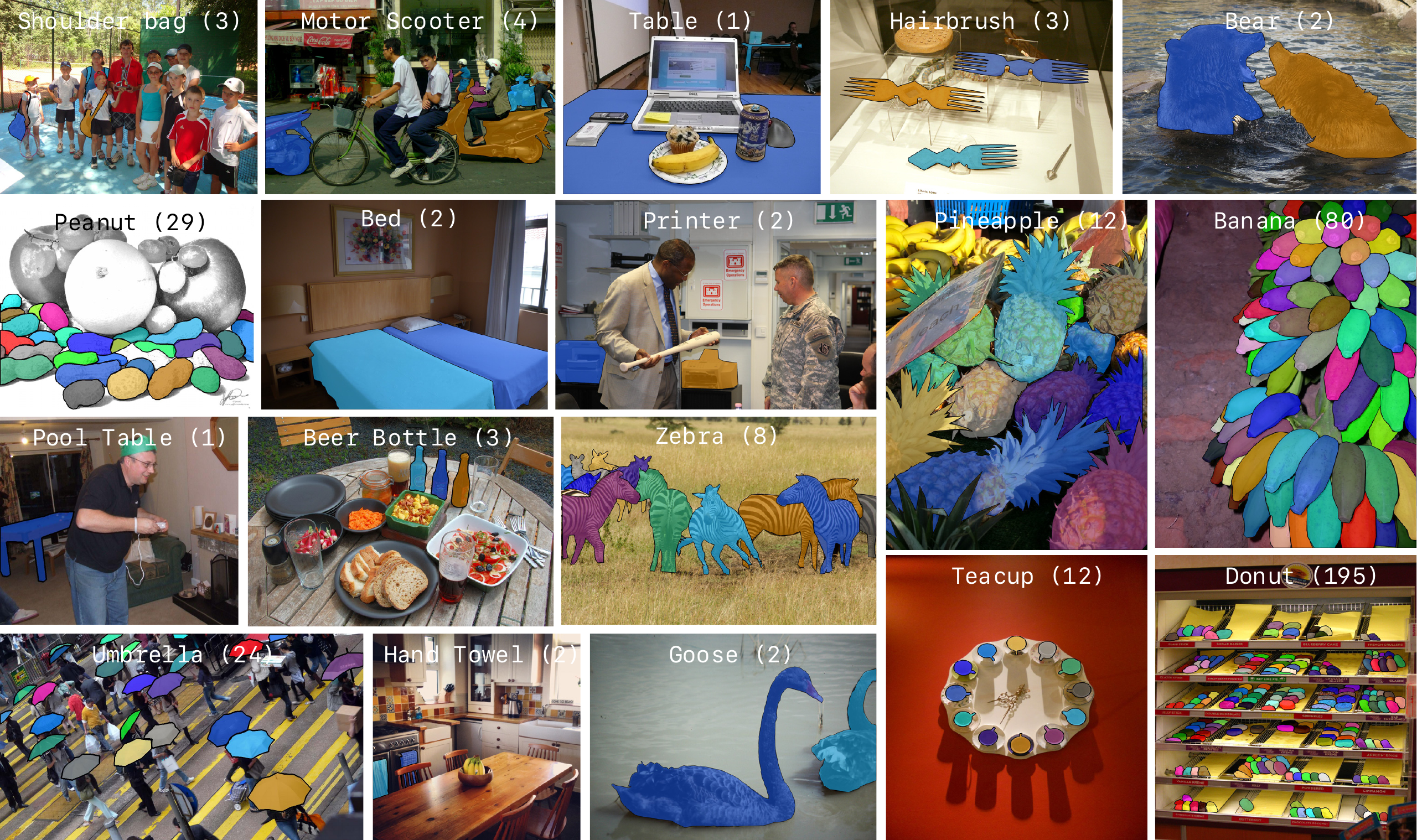}
  \caption{\textbf{Example \lvis annotations} (one category per image for clarity). See \url{http://www.lvisdataset.org/explore}.}
\label{fig:examples}
\vspace{-3mm}
\end{figure*}
%##################################################################################################

\subsection{Related Datasets}

Datasets shape the technical problems researchers study and consequently the path of scientific discovery~\cite{liberman2015}. We owe much of our current success in image recognition to pioneering datasets such as MNIST~\cite{mnist}, BSDS~\cite{MartinFTM01}, Caltech 101~\cite{fei2006one}, PASCAL VOC~\cite{Everingham2010}, ImageNet~\cite{Russakovsky2015}, and COCO~\cite{Lin2014}. These datasets enabled the development of algorithms that detect edges, perform large-scale image classification, and localize objects by bounding boxes and segmentation masks. They were also used in the discovery of important ideas, such as Convolutional Networks~\cite{LeCun1989,Krizhevsky2012}, Residual Networks~\cite{He2016}, and Batch Normalization~\cite{Ioffe2015}.

\lvis is inspired by these and other related datasets, including those focused on street scenes (Cityscapes~\cite{Cordts2016} and Mapillary~\cite{neuhold2017mapillary}) and pedestrians (Caltech Pedestrians~\cite{Dollar2012PAMI}). We review the most closely related datasets below.

\paragraph{COCO~\cite{Lin2014}} is the most popular instance segmentation benchmark for common objects. It contains 80 categories that are pairwise distinct. There are a total of 118k training images, 5k validation images, and 41k test images. All 80 categories are exhaustively annotated in all images (ignoring annotation errors), leading to approximately 1.2 million instance segmentation masks. To establish continuity with COCO, we adopt the same instance segmentation task and AP metric, and we are also annotating all images from the COCO 2017 dataset. All 80 COCO categories can be mapped into our dataset. In addition to representing an order of magnitude more categories than COCO, our annotation pipeline leads to higher-quality segmentation masks that more closely follow object boundaries (see \S\ref{sec:analysis}).

\paragraph{ADE20K~\cite{zhou2016semantic}} is an ambitious effort to annotate almost every pixel in 25k images with object instance, `stuff', and part segmentations. The dataset includes approximately 3000 named objects, stuff regions, and parts. Notably, ADE20K was annotated by a \emph{single expert annotator}, which increases consistency but also limits dataset size. Due to the relatively small number of annotated images, most of the categories do not have enough data to allow for both training and evaluation. Consequently, the instance segmentation benchmark associated with ADE20K evaluates algorithms on the 100 most frequent categories. In contrast, our goal is to enable benchmarking of \emph{large vocabulary} instance segmentation methods.

\paragraph{iNaturalist~\cite{vanhorn2017}} contains nearly 900k images annotated with bounding boxes for 5000 plant and animal species. Similar to our goals, iNaturalist emphasizes the importance of benchmarking classification and detection in the few example regime. Unlike our effort, iNaturalist does not include segmentation masks and is focussed on a different image and \emph{fine-grained} category distribution; our category distribution emphasizes entry-level categories.

\paragraph{Open Images v4~\cite{openimages}} is a large dataset of 1.9M images. The detection portion of the dataset includes 15M bounding boxes labeled with 600 object categories. The associated benchmark evaluates the 500 most frequent categories, all of which have over 100 training samples ($>$70\% of them have over 1000 training samples). Thus, unlike our benchmark, low-shot learning is not integral to Open Images. Also different from our dataset is the use of machine learning algorithms to select which images will be annotated by using classifiers for the target categories. Our data collection process, in contrast, involves no machine learning algorithms and instead discovers the objects that appear within a given set of images. Starting with release v4, Open Images has used a federated dataset design for object detection.

%%%%%%%%%%%%%%%%%%%%%%%%%%%%%%%%%%%%%%%%%%%%%%%%%%%%%%%%%%%%%%%%%%%%%%%%%%%%%%%%%%%%%%%%%%%%%%%%%%%
\section{Dataset Design}\label{sec:design}
We followed an \emph{evaluation-first design principle}: prior to any data collection, we precisely defined what task would be performed and how it would be evaluated. This principle is important because there are technical challenges that arise when evaluating detectors on a large vocabulary dataset that do not occur when there are few categories. These must be resolved first, because they have profound implications for the structure of the dataset, as we discuss next.

\subsection{Task and Evaluation Overview}
\paragraph{Task and Metric.} Our dataset benchmark is the instance segmentation task: given a fixed, known set of categories, design an algorithm that when presented with a previously unseen image will output a segmentation mask for each instance of each category that appears in the image along with the category label and a confidence score. Given the output of an algorithm over a set of images, we compute \emph{mask average precision} (AP) using the definition and implementation from the COCO dataset~\cite{cocoap} (for more detail see \S\ref{sec:evaldetails}).

\paragraph{Evaluation Challenges.} Datasets like PASCAL VOC and COCO use manually selected categories that are \emph{pairwise disjoint}: when annotating a \emph{car}, there's never any question if the object is instead a \emph{potted plant} or a \emph{sofa}. When increasing the number of categories, it is inevitable that other types of pairwise relationships will occur: (1) partially overlapping visual concepts; (2) parent-child relationships; and (3) perfect synonyms. See Fig.~\ref{fig:category_venn} for examples.

If these relations are not properly addressed, then the evaluation protocol will be unfair. For example, most \emph{toys} are not \emph{deer} and most \emph{deer} are not \emph{toys}, but a \emph{toy deer} is both---if a detector outputs \emph{deer} and the object is only labeled \emph{toy}, the detection will be marked as wrong. Likewise, if a car is only labeled \emph{vehicle}, and the algorithm outputs \emph{car}, it will be incorrectly judged to be wrong. Or, if an object is only labeled \emph{backpack} and the algorithm outputs the synonym \emph{rucksack}, it will be incorrectly penalized. Providing a fair benchmark is important for accurately reflecting algorithm performance.

These problems occur when the ground-truth annotations are missing one or more true labels for an object. If an algorithm happens to predict one of these correct, \emph{but missing} labels, it will be unfairly penalized. Now, if all objects are exhaustively and correctly labeled with all categories, then the problem is trivially solved. But correctly and exhaustively labeling 164k images each with 1000 categories is undesirable: it forces a binary judgement deciding if each category applies to each object; there will be many cases of genuine ambiguity and inter-annotator disagreement. Moreover, the annotation workload will be very large. Given these drawbacks, we describe our solution next.

\subsection{Federated Datasets}
Our key observation is that the desired evaluation protocol does not require us to exhaustively annotate all images with all categories. What is required instead is that for each category $c$ there must exist two disjoint subsets of the entire dataset $\allimgs$ for which the following guarantees hold:

\textbf{Positive set:} there exists a subset of images $\posc \subseteq \allimgs$ such that all instances of $c$ in $\posc$ are segmented. In other words, $\posc$ is exhaustively annotated for category $c$.

\textbf{Negative set:} there exists a subset of images $\negc \subseteq \allimgs$ such that no instance of $c$ appears in any of these images.

Given these two subsets for a category $c$, $\posc \cup \negc$ can be used to perform standard COCO-style AP evaluation for $c$. The evaluation oracle only judges the algorithm on a category $c$ over the subset of images in which $c$ has been exhaustively annotated; if a detector reports a detection of category $c$ on an image $\img \notin \posc \cup \negc$, the detection is \emph{not} evaluated.

By collecting the per-category sets into a single dataset, $\allimgs = \cup_{c} (\posc \cup \negc)$, we arrive at the concept of a \emph{federated dataset}. A federated dataset is a dataset that is formed by the union of smaller constituent datasets, each of which looks exactly like a traditional object detection dataset for a single category. By not annotating all images with all categories, freedom is created to design an annotation process that avoids ambiguous cases and collects annotations only if there is sufficient inter-annotator agreement. At the same time, the workload can be dramatically reduced.

Finally, we note that positive set and negative set membership on the test split is not disclosed and therefore algorithms have no side information about what categories will be evaluated in each image. An algorithm thus must make its best prediction for \emph{all} categories in each test image.

%##################################################################################################
\begin{figure*}[t]
\begin{center}
\includegraphics[width=1.0\linewidth]{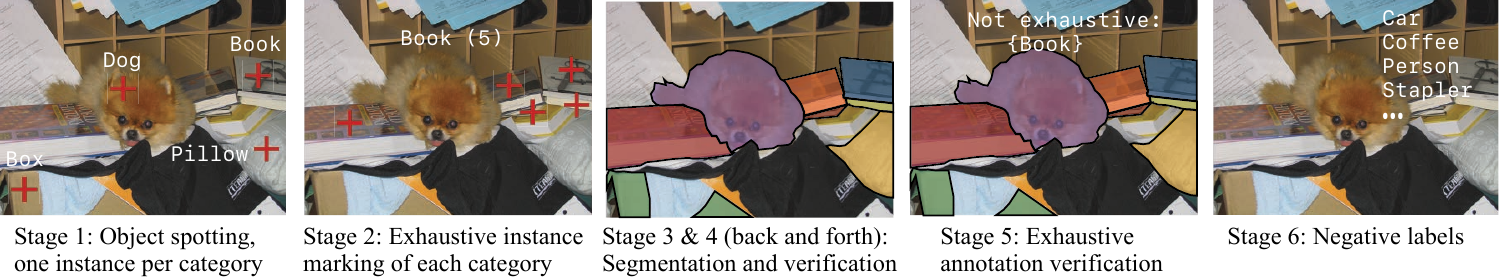}
\end{center}
\vspace{-2mm}
\caption{Our \textbf{annotation pipeline} comprises six stages. \textbf{Stage 1: Object Spotting} elicits annotators to mark a single instance of many different categories per image. This stage is iterative and causes annotators to discover a long tail of categories. \textbf{Stage 2: Exhaustive Instance Marking} extends the stage 1 annotations to cover all instances of each spotted category. Here we show additional instances of \emph{book}. \textbf{Stages 3 and 4: Instance Segmentation and Verification} are repeated back and forth until $\app$99\% of all segmentations pass a quality check. \textbf{Stage 5: Exhaustive Annotations Verification} checks that all instances are in fact segmented and flags categories that are missing one or more instances. \textbf{Stage 6: Negative Labels} are assigned by verifying that a subset of categories do not appear in the image.}
\label{fig:pipeline}
\vspace{-3mm}
\end{figure*}
%##################################################################################################

\paragraph{Reduced Workload.} Federated dataset design allows us to make $|\posc \cup \negc| \ll |\allimgs|, \forall c$. This choice dramatically reduces the workload and allows us to undersample the most frequent categories in order to avoid wasting annotation resources on them (\eg \emph{person} accounts for 30\% of COCO). Of our estimated \app 2 million instances, likely no single category will account for more than \app 3\% of the total instances.

\subsection{Evaluation Details}\label{sec:evaldetails}
The challenge evaluation server will only return the overall AP, not per-category AP's. We do this because: (1) it avoids leaking which categories are present in the \test set;\footnote{It's possible that the categories present in the \val and \test sets may be a strict subset of those in the \train set; we use the standard COCO 2017 \val and \test splits and cannot guarantee that all categories present in the \train images are also present in \val and \test.} (2) given that tail categories are rare, there will be few examples for evaluation in some cases, which makes per-category AP unstable; (3) by averaging over a large number of categories, the overall category-averaged AP has lower variance, making it a robust metric for ranking algorithms.

\paragraph{Non-Exhaustive Annotations.} We also collect an image-level boolean label, $\exhaustive$, indicating if image $\img \in \posc$ is exhaustively annotated for category $c$. In most cases (91\%), this flag is true, indicating that the annotations are indeed exhaustive. In the remaining cases, there is at least one instance in the image that is not annotated. Missing annotations often occur in `crowds' where there are a large number of instances and delineating them is difficult. During evaluation, we do not count false positives for category $c$ on images $\img$ that have $\exhaustive$ set to false. We do measure recall on these images: the detector is expected to predict accurate segmentation masks for the labeled instances. Our strategy differs from other datasets that use a small maximum number of instances per image, per category (10-15) together with `crowd regions' (COCO) or use a special `group of $c$' label to represent 5 or more instances (Open Images v4). Our annotation pipeline (\S\ref{sec:construction}) attempts to collect segmentations for \emph{all} instances in an image, regardless of count, and then checks if the labeling is in fact exhaustive. See Fig.~\ref{fig:examples}.

\paragraph{Hierarchy.}
During evaluation, we treat all categories the same; we do nothing special in the case of hierarchical relationships. To perform best, for each detected object $o$, the detector should output the most specific correct category as well as all more general categories, \eg, a canoe should be labeled both \emph{canoe} and \emph{boat}. The detected object $o$ in image $\img$ will be evaluated with respect to all labeled positive categories $\{c\ |\ \img \in \posc\}$, which may be any subset of categories between the most specific and the most general.

\paragraph{Synonyms.} A federated dataset that separates synonyms into different categories is valid, but is unnecessarily fragmented (see Fig.~\ref{fig:category_venn}, right). We avoid splitting synonyms into separate categories with WordNet~\cite{miller1998wordnet}. Specifically, in \lvis each category $c$ is a WordNet \emph{synset}---a word sense specified by a set of synonyms and a definition.

%%%%%%%%%%%%%%%%%%%%%%%%%%%%%%%%%%%%%%%%%%%%%%%%%%%%%%%%%%%%%%%%%%%%%%%%%%%%%%%%%%%%%%%%%%%%%%%%%%%
\section{Dataset Construction}\label{sec:construction}
In this section we provide an overview of the annotation pipeline that we use to collect \lvis.

\subsection{Annotation Pipeline}
Fig.~\ref{fig:pipeline} illustrates our annotation pipeline by showing the output of each stage, which we describe below. For now, assume that we have a fixed category vocabulary $\vocab$. We will describe how the vocabulary was collected in \S\ref{sec:vocab}.

\paragraph{Object Spotting, Stage 1.} The goals of the object spotting stage are to: (1) generate the positive set, $\posc$, for each category $c \in \vocab$ and (2) elicit vocabulary recall such that many different object categories are included in the dataset.

Object spotting is an iterative process in which each image is visited a variable number of times. On the first visit, an annotator is asked to mark one object with a point and to name it with a category $c \in \vocab$ using an \emph{autocomplete} text input. On each subsequent visit, all previously spotted objects are displayed and an annotator is asked to mark an object of a previously unmarked category or to skip the image if no more categories in $\vocab$ can be spotted. When an image has been skipped 3 times, it will no longer be visited. The autocomplete is performed against the set of all synonyms, presented with their definitions; we internally map the selected word to its synset/category to resolve synonyms.

Obvious and salient objects are spotted early in this iterative process. As an image is visited more, less obvious objects are spotted, including incidental, non-salient ones. We run the spotting stage twice, and for each image we retain categories that were spotted in both runs. \emph{Thus two people must independently agree on a name in order for it to be included in the dataset; this increases naming consistency.}

To summarize the output of stage 1: for each category in the vocabulary, we have a (possibly empty) set of images in which one object of that category is marked per image. This defines an initial positive set, $\posc$, for each category $c$.

\paragraph{Exhaustive Instance Marking, Stage 2.} The goals this stage are to: (1) verify stage 1 annotations and (2) take each image $\img \in \posc$ and mark \emph{all} instances of $c$ in $\img$ with a point.

In this stage, $(\img, c)$ pairs from stage 1 are each sent to 5 annotators. They are asked to perform two steps. First, they are shown the definition of category $c$ and asked to verify if it describes the spotted object. Second, if it matches, then the annotators are asked to mark all other instances of the same category. If it does not match, there is no second step. To prevent frequent categories from dominating the dataset and to reduce the overall workload, we subsample frequent categories such that no positive set exceeds more than 1\% of the images in the dataset.

To ensure annotation quality, we embed a `gold set' within the pool of work. These are cases for which we know the correct ground-truth. We use the gold set to automatically evaluate the work quality of each annotator so that we can direct work towards more reliable annotators. We use 5 annotators per $(\img, c)$ pair to help ensure instance-level recall.

To summarize, from stage 2 we have exhaustive instance spotting for each image $\img \in \posc$ for each category $c \in \vocab$.

\paragraph{Instance Segmentation, Stage 3.} The goals of the instance segmentation stage are to: (1) verify the category for each marked object from stage 2 and (2) upgrade each marked object from a point annotation to a full segmentation mask.

To do this, each pair $(\img, o)$ of image $\img$ and marked object instance $o$ is presented to one annotator who is asked to verify that the category label for $o$ is correct and if it is correct, to draw a \emph{detailed} segmentation mask for it (\eg see Fig.~\ref{fig:examples}).

We use a training task to establish our quality standards. Annotator quality is assessed with a gold set and by tracking their average vertex count per polygon. We use these metrics to assign work to reliable annotators.

In sum, from stage 3 we have for each image and spotted instance pair one segmentation mask (if it is not rejected).

\paragraph{Segment Verification, Stage 4.} The goal of the segment verification stage is to verify the quality of the segmentation masks from stage 3. We show each segmentation to up to 5 annotators and ask them to rate its quality using a rubric. If two or more annotators reject the mask, then we requeue the instance for stage 3 segmentation. Thus we only accept a segmentation if 4 annotators agree it is high-quality. Unreliable workers from stage 3 are not invited to judge segmentations in stage 4; we also use rejections rates from this stage to monitor annotator reliability. We iterate between stages 3 \& 4 a total of four times, each time only re-annotating rejected instances.

To summarize the output of stage 4 (after iterating back and forth with stage 3): we have a high-quality segmentation mask for $>$99\% of all marked objects.

\paragraph{Full Recall Verification, Stage 5.} The full recall verification stage finalizes the positive sets. The goal is to find images $\img \in \posc$ where $c$ is not exhaustively annotated. We do this by asking annotators if there are any unsegmented instances of category $c$ in $\img$. We ask up to 5 annotators and require at least 4 to agree that annotation is exhaustive. As soon as two believe it is not, we mark the exhaustive annotation flag $\exhaustive$ as false. We use a gold set to maintain quality.

To summarize the output of stage 5: we have a boolean flag $\exhaustive$ for each image $\img \in \posc$ indicating if category $c$ is exhaustively annotated in image $\img$. This finalizes the positive sets along with their instance segmentation annotations.

\paragraph{Negative Sets, Stage 6.} The final stage of the pipeline is to collect a negative set $\negc$ for each category $c$ in the vocabulary. We do this by randomly sampling images $\img \in \allimgs \setminus \posc$, where $\allimgs$ is all images in the dataset. For each sampled image $\img$, we ask up to 5 annotators if category $c$ appears in image $\img$. If any one annotator reports that it does, we reject the image. Otherwise $\img$ is added to $\negc$. We sample until the negative set $\negc$ reaches a target size of 1\% of the images in the dataset. We use a gold set to maintain quality.

To summarize, from stage 6 we have a negative image set $\negc$ for each category $c \in \vocab$ such that the category does not appear in any of the images in $\negc$.

\subsection{Vocabulary Construction}\label{sec:vocab}
We construct the vocabulary $\vocab$ with an iterative process that starts from a large super-vocabulary and uses the object spotting process (stage 1) to winnow it down. We start from 8.8k synsets that were selected from WordNet by removing some obvious cases (\eg proper nouns) and then finding the intersection with highly concrete common nouns~\cite{brysbaert2014concreteness}. This yields a high-recall set of concrete, and thus likely visual, entry-level synsets. We then apply object spotting to 10k COCO images with autocomplete against this super-vocabulary. This yields a reduced vocabulary with which we repeat the process once more. Finally, we perform minor manual editing. The resulting vocabulary contains 1723 synsets---the upper bound on the number of categories that can appear in \lvis.

%##################################################################################################
\begin{figure}[t]\centering
\includegraphics[width=0.99\linewidth]{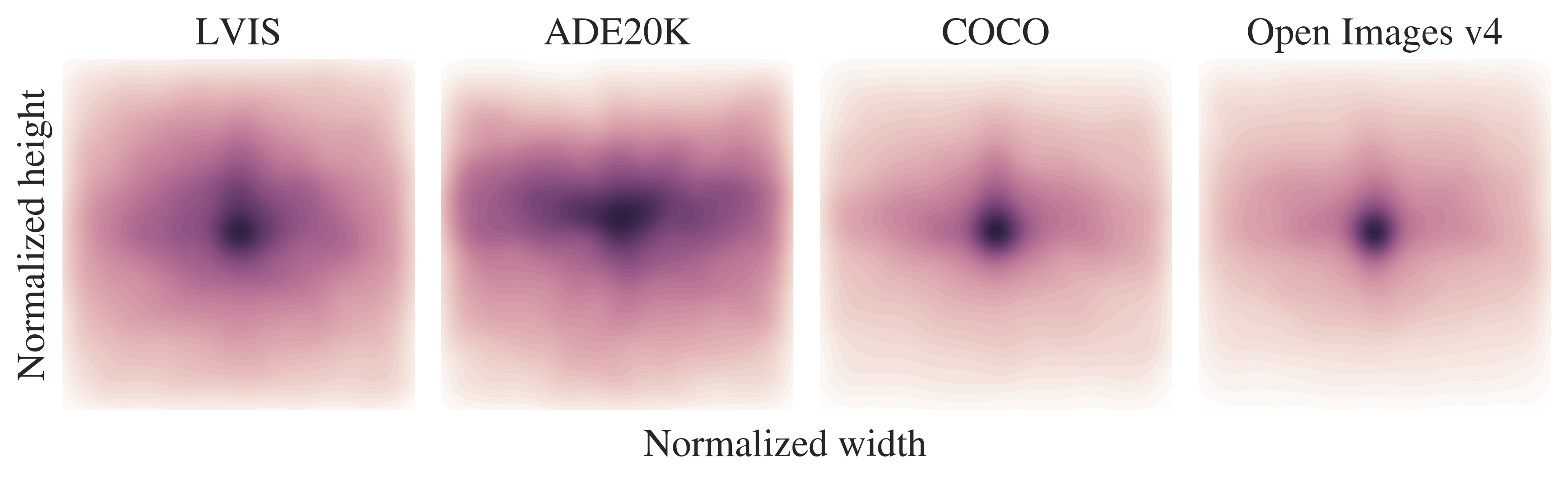}
  \caption{Distribution of object centers in normalized image coordinates for four datasets. ADE20K exhibits the greatest spatial diversity, with \lvis achieving greater complexity than COCO and the Open Images v4 training set.\textsuperscript{\ref{oidplot}}}
\label{fig:analysis:center_distribution}
\vspace{-3mm}
\end{figure}
%##################################################################################################

%##################################################################################################
\begin{figure*}[t]\centering
  \subfloat[\label{fig:analysis:percent_category_count}Distribution of category count per image. \lvis has a heavier tail than COCO and Open Images training set. ADE20K is the most uniform.]{\includegraphics[width=0.32\linewidth]{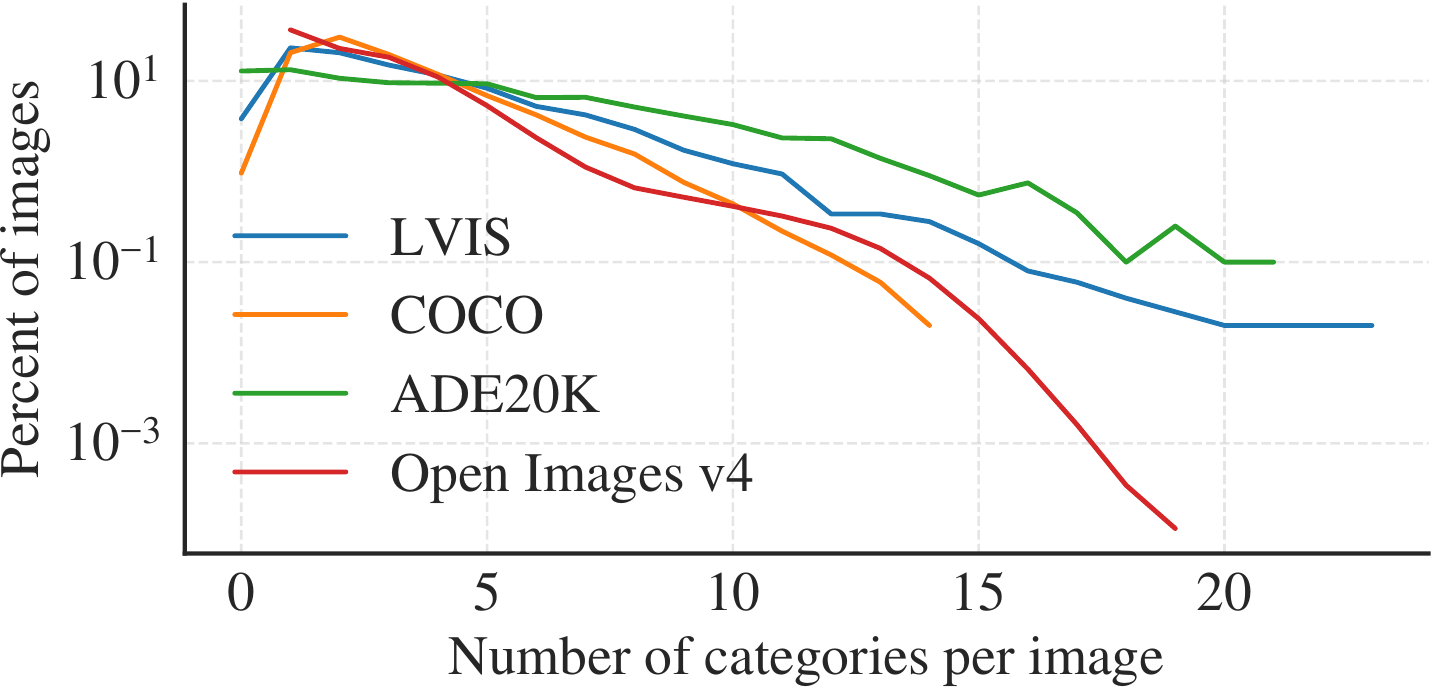}}
\hfill
\subfloat[\label{fig:analysis:instances_per_category}The number of instances per category (on 5k images) reveals the long tail with few examples. Orange dots: categories in common with COCO.]{\includegraphics[width=0.32\linewidth]{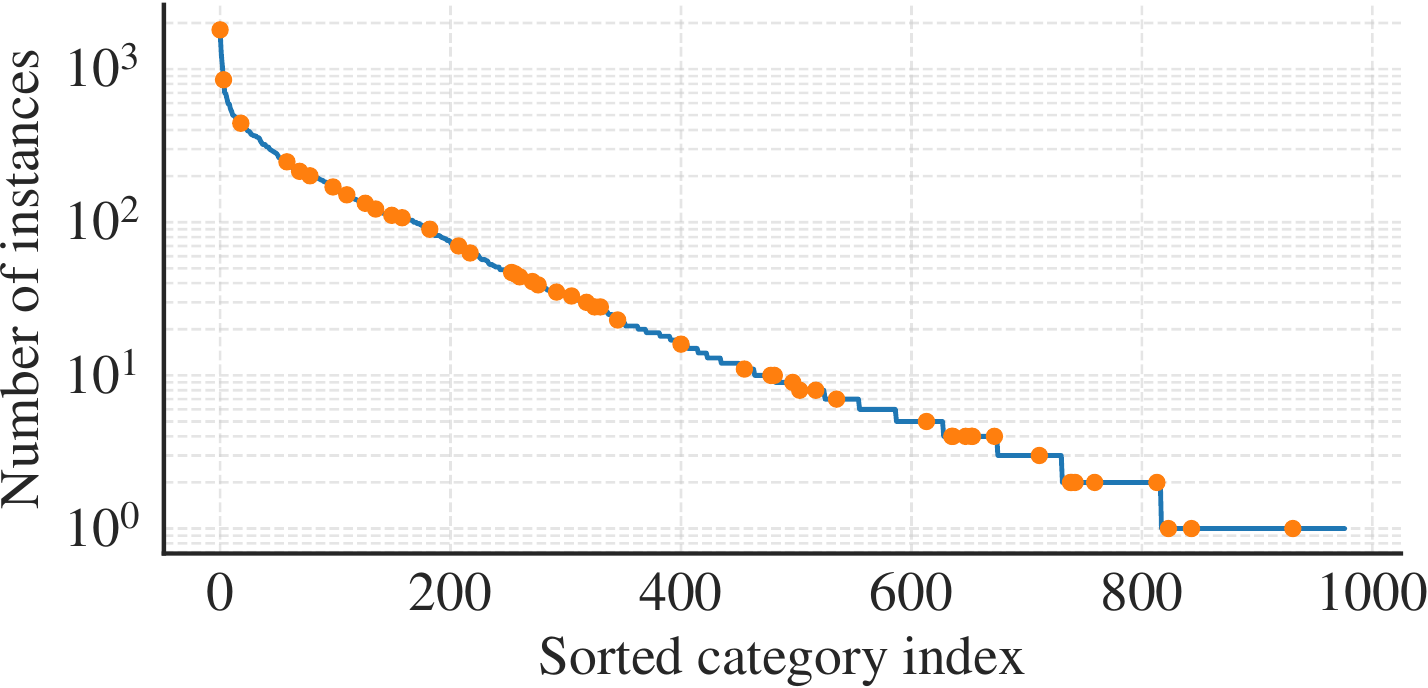}}
\hfill
\subfloat[\label{fig:analysis:area_distribution}Relative segmentation mask size (square root of mask-area-divided-by-image-area) compared between \lvis, COCO, and ADE20K.]{\includegraphics[width=0.32\linewidth]{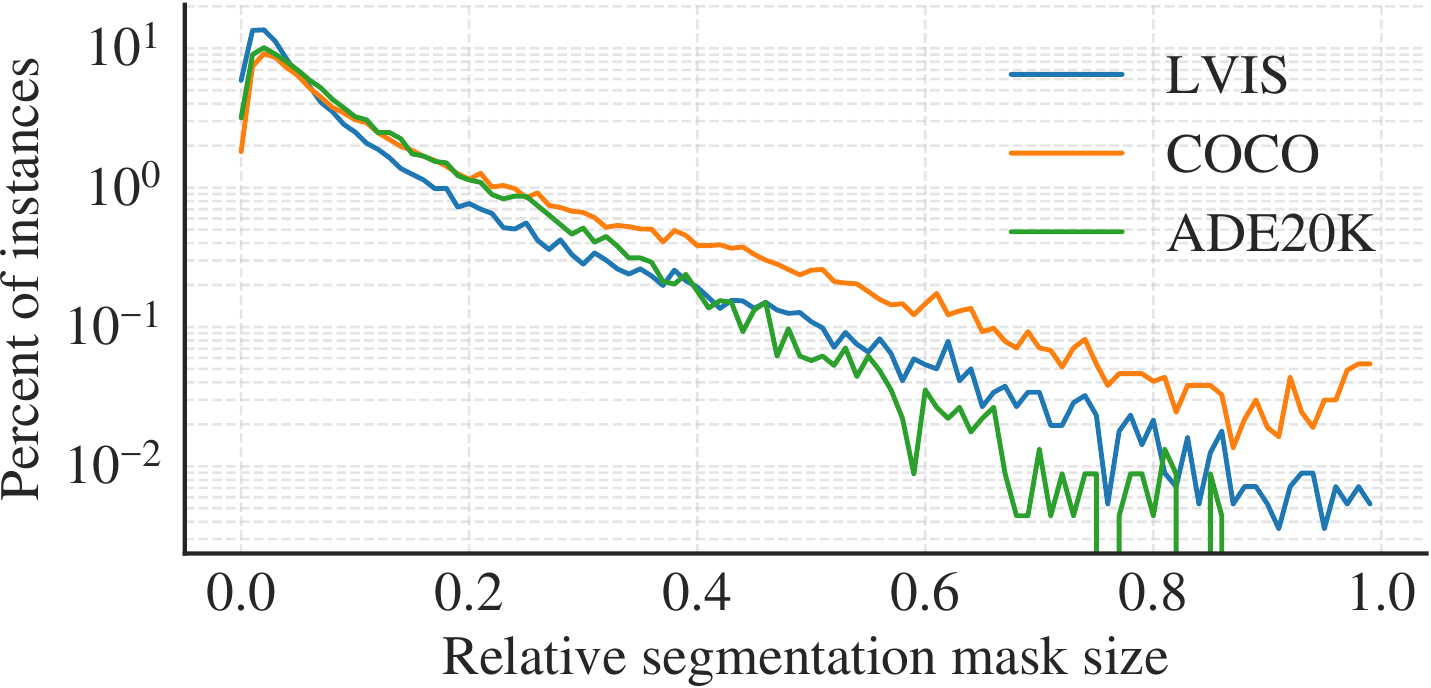}}\\[.25em]
\caption{\textbf{Dataset statistics}. Best viewed digitally.}
\vspace{-3mm}
\end{figure*}
%##################################################################################################

%##################################################################################################
\begin{figure*}[!h]\centering
\subfloat[\label{fig:analysis:sq}\lvis segmentation quality measured by mask IoU between matched instances from two runs of our annotation pipeline. Masks from the runs are consistent with a dataset average IoU of 0.85.]{\includegraphics[width=0.32\linewidth]{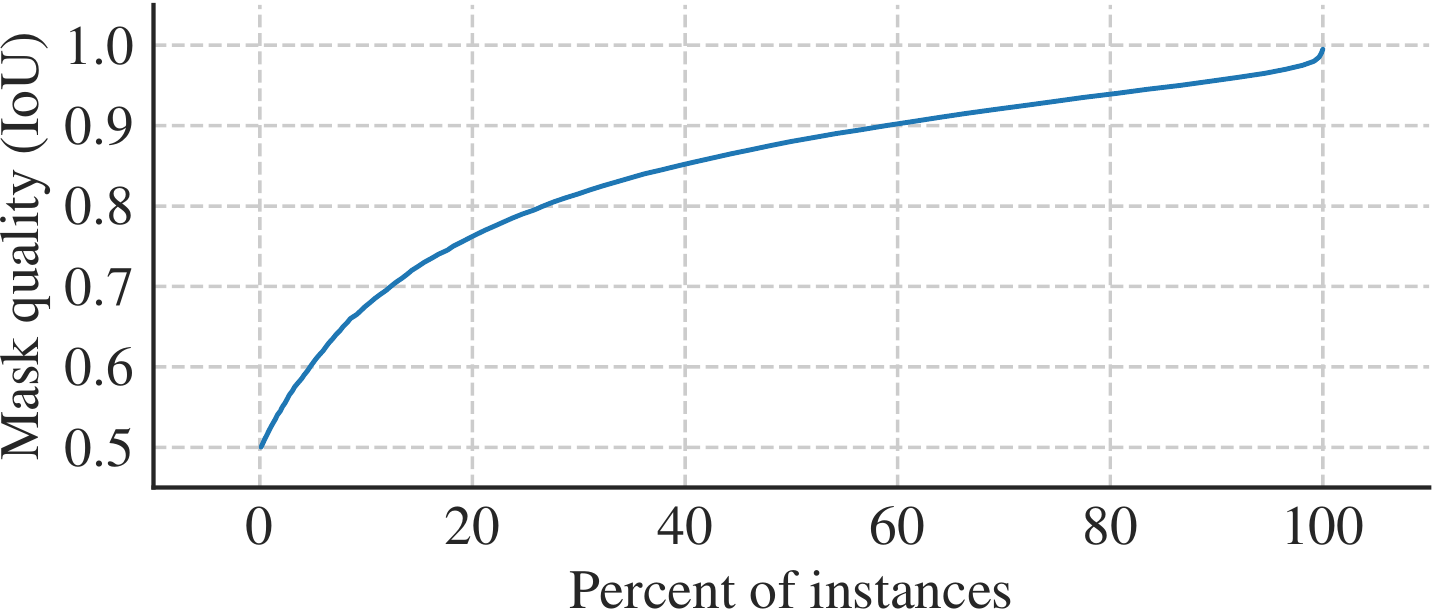}}
\hfill
\subfloat[\label{fig:analysis:rq}\lvis recognition quality measured by $F_1$ score given matched instances across two runs of our annotation pipeline. Category labeling is consistent with a dataset average $F_1$ score of 0.87.]{\includegraphics[width=0.32\linewidth]{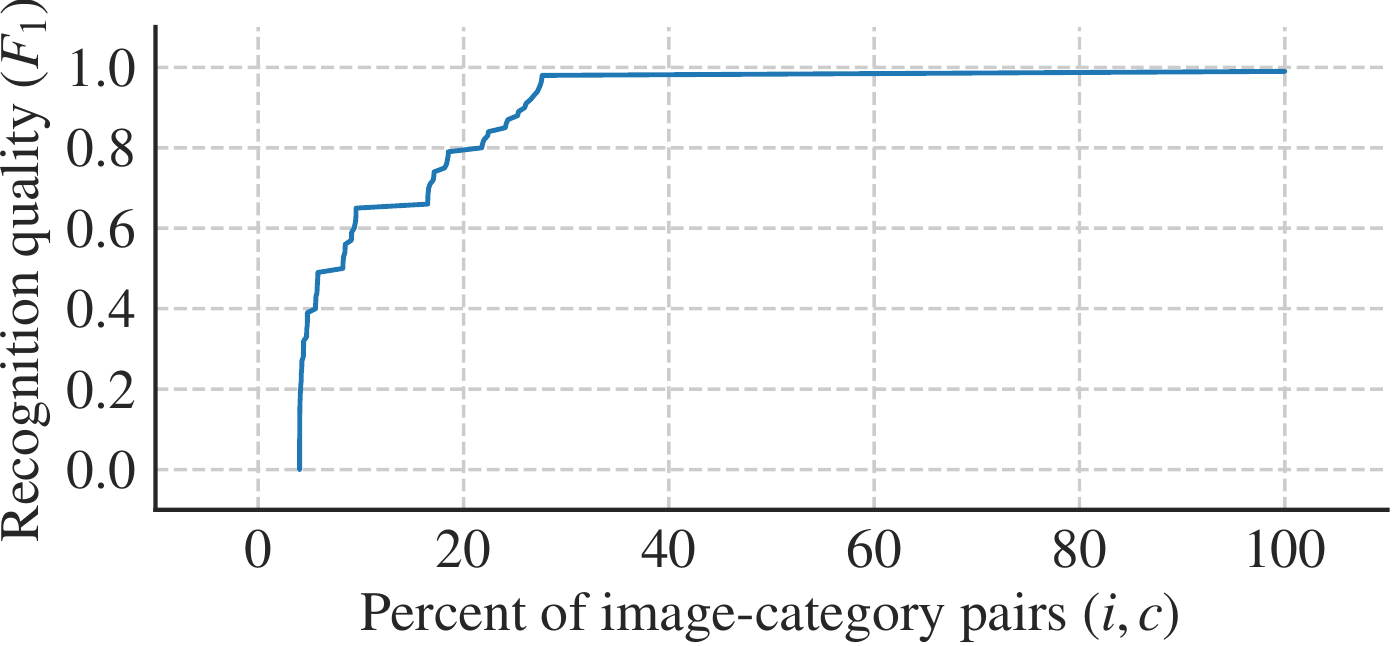}}
\hfill
\subfloat[\label{fig:analysis:mask}Illustration of mask IoU \vs boundary quality to provide intuition for interpreting Fig.~\ref{fig:analysis:sq} (left) and Tab.~\ref{tab:analysis:expert} (dataset annotations \vs expert annotators, below).]{\includegraphics[width=0.32\linewidth]{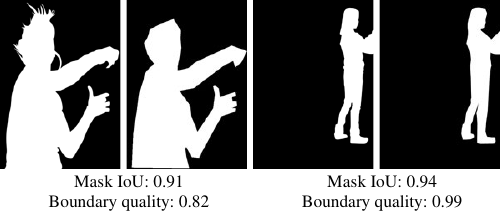}}\\[.25em]
\caption{\textbf{Annotation consistency} using 5000 \emph{doubly annotated} images from \lvis. Best viewed digitally.}
\vspace{-3mm}
\end{figure*}
%##################################################################################################

%##################################################################################################
\begin{table*}[!h]\centering
% subfloat a - annotation quality
\subfloat[\label{tab:analysis:expert}For each metric (mask IoU, boundary quality) and each statistic (mean, median), we show a bootstrapped 95\% confidence interval. \lvis has the highest quality across all measures.]{%
\tablestyle{6.5pt}{1.0}\begin{tabular}{@{}ll|cc|cc@{}}
& & \multicolumn{2}{c|}{mask IoU} & \multicolumn{2}{c}{boundary quality} \\
dataset & comparison & mean & median & mean & median \\
\shline
\multirow{2}{*}{COCO} & dataset \vs experts & 0.83 \h 0.87 & 0.88 \h 0.91 & 0.77 \h 0.82 & 0.79 \h 0.88 \\
 & \demph{\eve} & \demph{0.91 \h 0.95} & \demph{0.96 \h 0.98} & \demph{0.92 \h 0.96} & \demph{0.97 \h 0.99} \\
\hline
\multirow{2}{*}{ADE20K} & dataset \vs experts & 0.84 \h 0.88 & 0.90 \h 0.93 & 0.83 \h 0.87 & 0.84 \h 0.92 \\
 & \demph{\eve} & \demph{0.90 \h 0.94} & \demph{0.95 \h 0.97} & \demph{0.90 \h 0.95} & \demph{0.99 \h 1.00} \\
\hline
\multirow{2}{*}{\lvis} & dataset \vs experts & \textbf{0.90 \h 0.92} & \textbf{0.94 \h 0.96} & \textbf{0.87 \h 0.91} & \textbf{0.93 \h 0.98} \\
 & \demph{\eve} & \demph{0.93 \h 0.96} & \demph{0.96 \h 0.98} & \demph{0.91 \h 0.96} & \demph{0.97 \h 1.00} \\
\end{tabular}}
\hspace{6mm}
% subfloat b - annotation complexity
\subfloat[\label{tab:analysis:complexity}Comparison of annotation complexity. Boundary complexity is perimeter divided by square root area~\cite{attneave1956quantitative}.]{%
\tablestyle{6.5pt}{1.0}\begin{tabular}{@{}lc|cc@{}}
 & annotation & \multicolumn{2}{c}{boundary complexity} \\
dataset\ \ \ & source & mean & median \\
\shline
\multirow{2}{*}{COCO} & dataset & 5.59 \h 6.04 & 5.13 \h 5.51 \\
 & \demph{experts} & \demph{6.94 \h 7.84} & \demph{5.86 \h 6.80} \\
\hline
\multirow{2}{*}{ADE20K} & dataset & 6.00 \h 6.84 & 4.79 \h 5.31 \\
 & \demph{experts} & \demph{6.34 \h 7.43} & \demph{4.83 \h 5.53} \\
\hline
\multirow{2}{*}{\lvis} & dataset & \textbf{6.35 \h 7.07} & \textbf{5.44 \h 6.00} \\
 & \demph{experts} & \demph{7.13 \h 8.48} & \demph{5.91 \h 6.82} \\
\end{tabular}}\\[.25em]
% main caption
\caption{Annotation \textbf{quality and complexity} relative to experts.}
\label{tab:analysis:experts}
\vspace{-3mm}
\end{table*}
%##################################################################################################

%%%%%%%%%%%%%%%%%%%%%%%%%%%%%%%%%%%%%%%%%%%%%%%%%%%%%%%%%%%%%%%%%%%%%%%%%%%%%%%%%%%%%%%%%%%%%%%%%%%
\section{Dataset Analysis}\label{sec:analysis}
For analysis, we have annotated 5000 images (the COCO \texttt{val2017} split) twice using the proposed pipeline. We begin by discussing general dataset statistics next before proceeding to an analysis of annotation consistency in \S\ref{sec:analyis:consistency} and an analysis of the evaluation protocol in \S\ref{sec:analyis:eval}.

\subsection{Dataset Statistics}\label{sec:analysis:stats}

\paragraph{Category Statistics.}
There are 977 categories present in the 5000 \lvis images. The category growth rate (see Fig.~\ref{fig:analysis:cumulative_category_count}) indicates that the final dataset will have well over 1000 categories. On average, each image is annotated with 11.2 instances from 3.4 categories. The largest instances-per-image count is a remarkable 294. Fig.~\ref{fig:analysis:percent_category_count} shows the full categories-per-image distribution. \lvis's distribution has more spread than COCO's indicating that many images are labeled with more categories. The low-shot nature of our dataset can be seen in Fig.~\ref{fig:analysis:instances_per_category}, which plots the total number of instances for each category (in the 5000 images). The median value is 9, and while this number will be larger for the full image set, this statistic highlights the challenging long-tailed nature of our data.

\paragraph{Spatial Statistics.}
Our object spotting process (stage 1) encourages the inclusion of objects distributed throughout the image plane, not just the most salient foreground objects. The effect can be seen in Fig.~\ref{fig:analysis:center_distribution} which shows object-center density plots. All datasets have some degree of center bias, with ADE20K and \lvis having the most diverse spatial distribution. COCO and Open Images v4 (training set\footnote{\label{oidplot}The CVPR 2019 version of this paper shows the distribution of the Open Images v4 \emph{validation} set, which has more center bias. The peakiness is also exaggerated due to an intensity scaling artifact. For more details, see \url{https://storage.googleapis.com/openimages/web/factsfigures.html}.}) have similar object-center distributions with a marginally lower degree of spatial diversity.

\paragraph{Scale Statistics.}
Objects in \lvis are also more likely to be small. Fig.~\ref{fig:analysis:area_distribution} shows the relative size distribution of object masks: compared with COCO, \lvis objects tend to smaller and there are fewer large objects (\eg, objects that occupy most of an image are $\app$10$\times$ less frequent). ADE20K has the fewest large objects overall and more medium ones.

\subsection{Annotation Consistency}\label{sec:analyis:consistency}

\paragraph{Annotation Pipeline Repeatability.}
A repeatable annotation pipeline implies that the process generating the ground-truth data is not overly random and therefore may be learned. To understand repeatability, we annotated the 5000 images twice: after completing object spotting (stage 1), we have initial positive sets $\posc$ for each category $c$; we then execute stages 2 through 5 (exhaustive instance marking through full recall verification) twice in order to yield doubly annotated positive sets. To compare them, we compute a matching between them for each image and category pair. We find a matching that maximizes the total mask intersection over union (IoU) summed over the matched pairs and then discard any matches with IoU $<$ 0.5. Given these matches we compute the dataset average mask IoU (0.85) and the dataset average $F_1$ score (0.87). Intuitively, these quantities describe `segmentation quality' and `recognition quality'~\cite{kirillov2018panoptic}. The cumulative distributions of these metrics (Fig.~\ref{fig:analysis:sq} and~\ref{fig:analysis:rq}) show that even though matches are established based on a low IoU threshold (0.5), matched masks tend to have much higher IoU. The results show that roughly 50\% of matched instances have IoU greater than 90\% and roughly 75\% of the image-category pairs have a perfect $F_1$ score. Taken together, these metrics are a strong indication that our pipeline has a large degree of repeatability.

\paragraph{Comparison with Expert Annotators.}
To measure segmentation quality, we randomly selected 100 instances with mask area greater than $32^2$ pixels from \lvis, COCO, and ADE20K. We presented these instances (indicated by bounding box and category) to two independent expert annotators and asked them to segment each object using professional image editing tools. We compare dataset annotations to expert annotations using mask IoU and boundary quality (boundary $F$~\cite{MartinFTM01}) in Tab.~\ref{tab:analysis:expert}. The results (bootstrapped 95\% confidence intervals) show that our masks are high-quality, surpassing COCO and ADE20K on both measures (see Fig.~\ref{fig:analysis:mask} for intuition). At the same time, the objects in \lvis have more complex boundaries~\cite{attneave1956quantitative} (Tab.~\ref{tab:analysis:complexity}).

%##################################################################################################
\begin{figure*}[t]\centering
\subfloat[\label{fig:analysis:neg}Given fixed detections, we show how AP varies with $|\negc|$, the number of negative images per category used in evaluation.]{\includegraphics[width=0.32\linewidth]{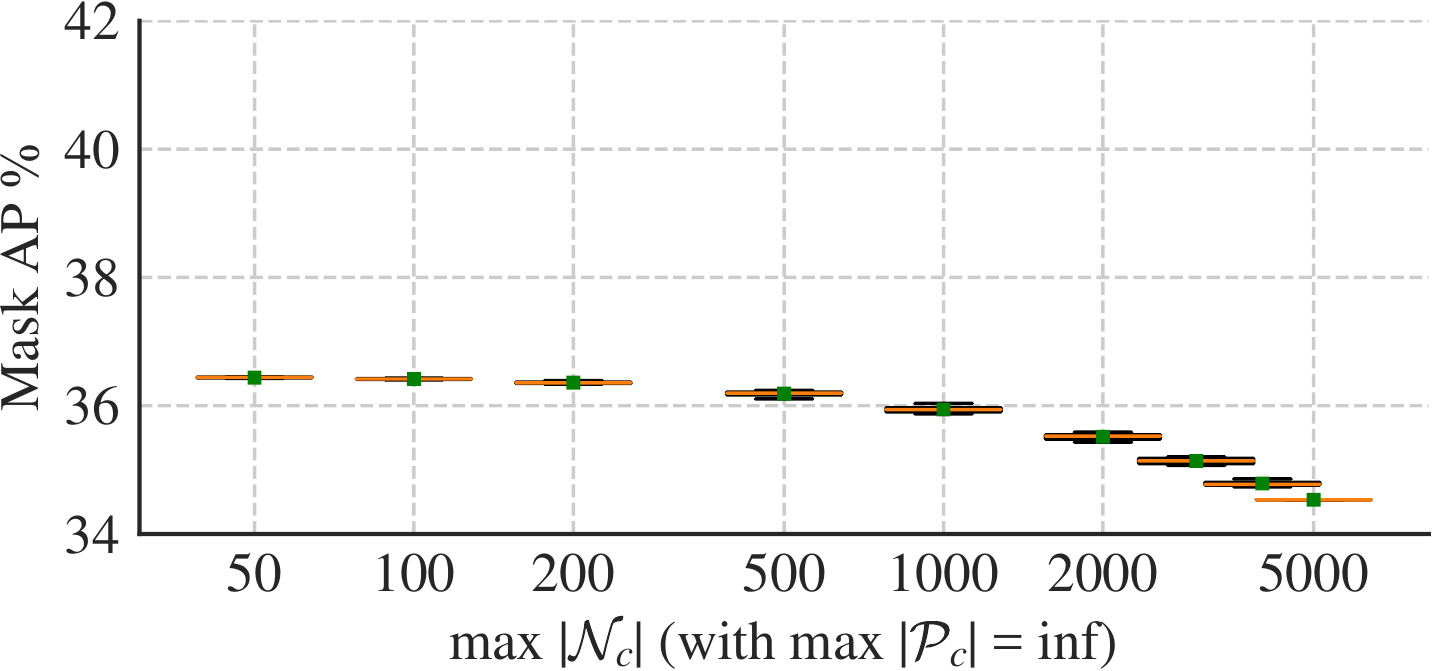}}
\hfill
\subfloat[\label{fig:analysis:pos}With the same detections from Fig.~\ref{fig:analysis:neg} and $|\negc| = 50$, we show how AP varies as we vary $|\posc|$, the positive set size.]{\includegraphics[width=0.32\linewidth]{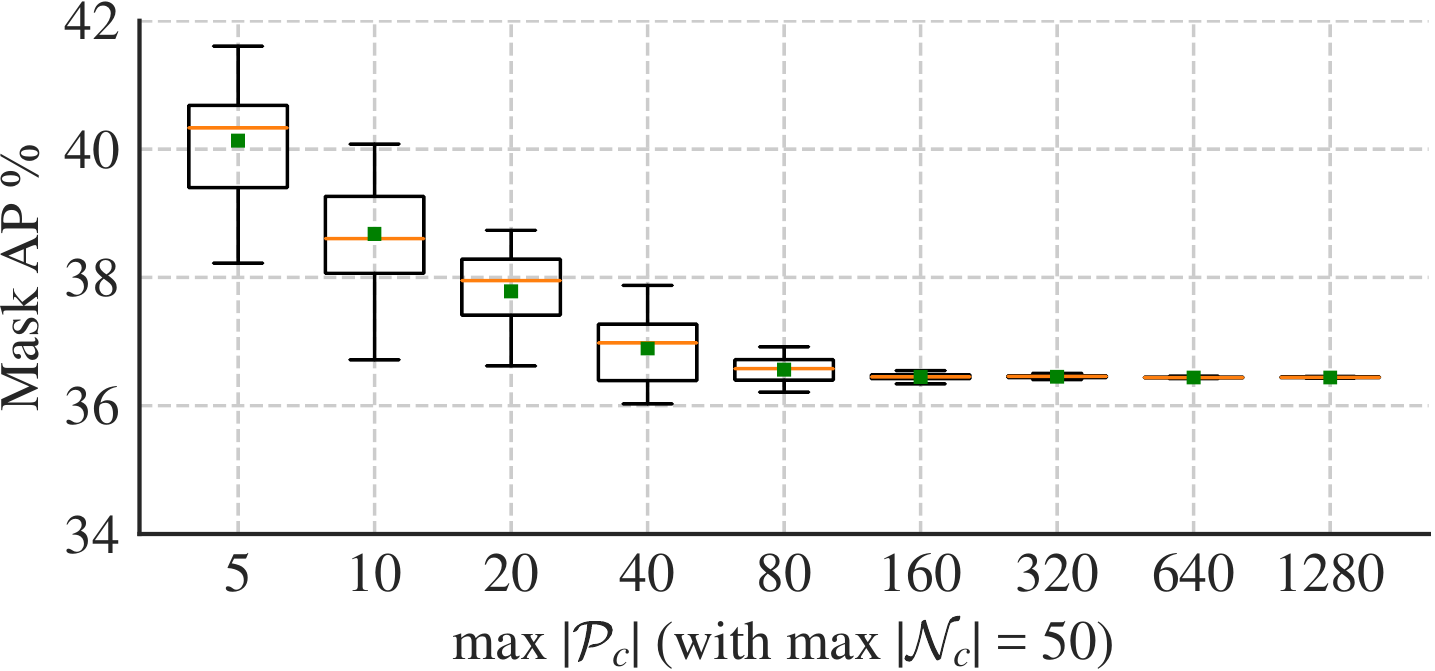}}
\hfill
\subfloat[\label{fig:analysis:low_shot}Low-shot detection is an open problem: training Mask R-CNN on 1k images decreases COCO \texttt{val2017} mask AP from 36\% to 10\%.]{\includegraphics[width=0.31\linewidth]{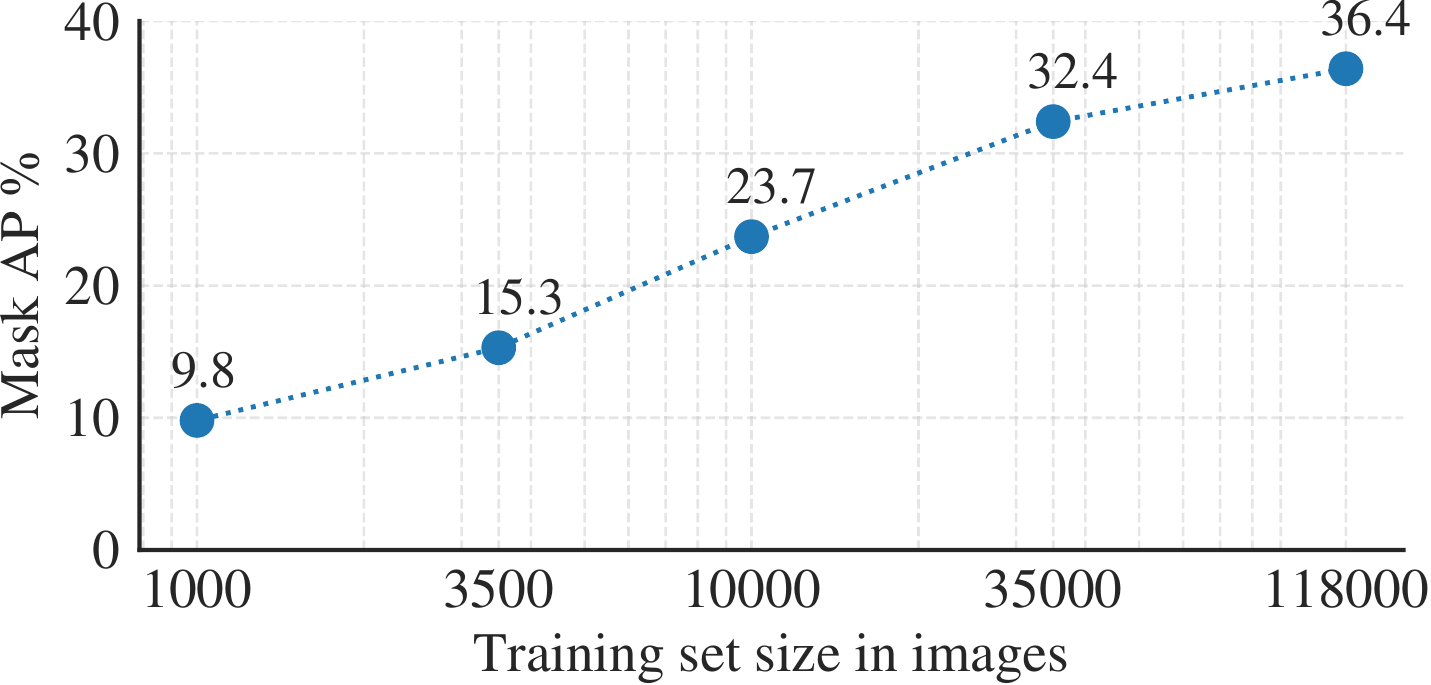}}\\[.25em]
  \caption{\textbf{Analysis of AP as a function of different data sizes.} Best viewed digitally.}
\vspace{-7mm}
\end{figure*}
%##################################################################################################

%##################################################################################################
\begin{table}[h]\centering
\tablestyle{8pt}{1.05}\begin{tabular}{ll|c|c}
\multicolumn{4}{c}{~} \\
Mask R-CNN & test anno. & box AP & mask AP \\
\shline
 ResNet-50-FPN & COCO & 38.2 & 34.1 \\
 \tiny\demph{model id: 35859007} & \lvis & 38.8 & 34.4 \\
\hline
 ResNet-101-FPN & COCO & 40.6 & 36.0 \\
 \tiny\demph{model id: 35861858} & \lvis & 40.9 & 36.0 \\
\hline
 ResNeXt-101-64x4d-FPN & COCO & 47.8 & 41.2 \\
 \tiny\demph{model id: 37129812} & \lvis & 48.6 & 41.7 \\
\end{tabular}\\[.25em]
\vspace{0.5mm}
\begin{minipage}{.92\linewidth}
\caption{COCO-trained Mask R-CNN evaluated on \lvis annotations. Both annotations yield similar AP values.}
\label{tab:analysis:detector_eval}
\end{minipage}
\vspace{-3mm}
\end{table}
%##################################################################################################

\subsection{Evaluation Protocol}\label{sec:analyis:eval}

\paragraph{COCO Detectors on \lvis.}
To validate our annotations and federated dataset design we downloaded three Mask R-CNN~\cite{He2017} models from the Detectron Model Zoo~\cite{Detectron2018} and evaluated them on \lvis annotations for the categories in COCO\@. Tab.~\ref{tab:analysis:detector_eval} shows that both box AP and mask AP are close between our annotations and the original ones from COCO for all models, which span a wide AP range. This result validates our annotations and evaluation protocol: even though \lvis uses a federated dataset design with sparse annotations, the quantitative outcome closely reproduces the `gold standard' results from dense COCO annotations.

\paragraph{Federated Dataset Simulations.} For insight into how AP changes with positive and negative sets sizes $|\posc|$ and $|\negc|$, we randomly sample smaller evaluation sets (20 times) from COCO \texttt{val2017} and recompute AP\@. In Fig.~\ref{fig:analysis:neg} we use all positive instances for evaluation, but vary $|\negc|$ between 50 and 5k. AP decreases somewhat ($\app$2\% absolute) as we increase the number of negative images as the ratio of negative to positive examples grows with fixed $|\posc|$ and increasing $|\negc|$. Next, in Fig.~\ref{fig:analysis:pos} we set $|\negc|=50$ and vary $|\posc|$. We observe that even with a small positive set size of 80, AP is similar to the baseline with low variance. With smaller positive sets (down to 5) variance increases, but the AP gap from 1st to 3rd quartile remains below 2\% absolute. A curious upward bias in AP appears, which we investigate in \S\ref{sec:appendix:ap_bias}. These simulations together with COCO detectors tested on \lvis (Tab.~\ref{tab:analysis:detector_eval}) indicate that including smaller evaluation sets for each category is viable for evaluation.

\paragraph{Low-Shot Detection.}
To validate the claim that low-shot detection is a challenging open problem, we trained Mask R-CNN on random subsets of COCO \texttt{train2017} ranging from 1k to 118k images. For each subset, we optimized the learning rate schedule and weight decay by grid search. Results on \texttt{val2017} are shown in Fig.~\ref{fig:analysis:low_shot}. At 1k images, mask AP drops from 36.4\% (full dataset) to 9.8\% (1k subset). In the 1k subset, 89\% of the categories have more than 20 training instances, while the low-shot literature typically considers $\ll$ 20 examples per category~\cite{hariharan2017lowshot}.

%##################################################################################################
\begin{figure}[t]\centering
\vspace{4mm}
\includegraphics[height=2.5cm]{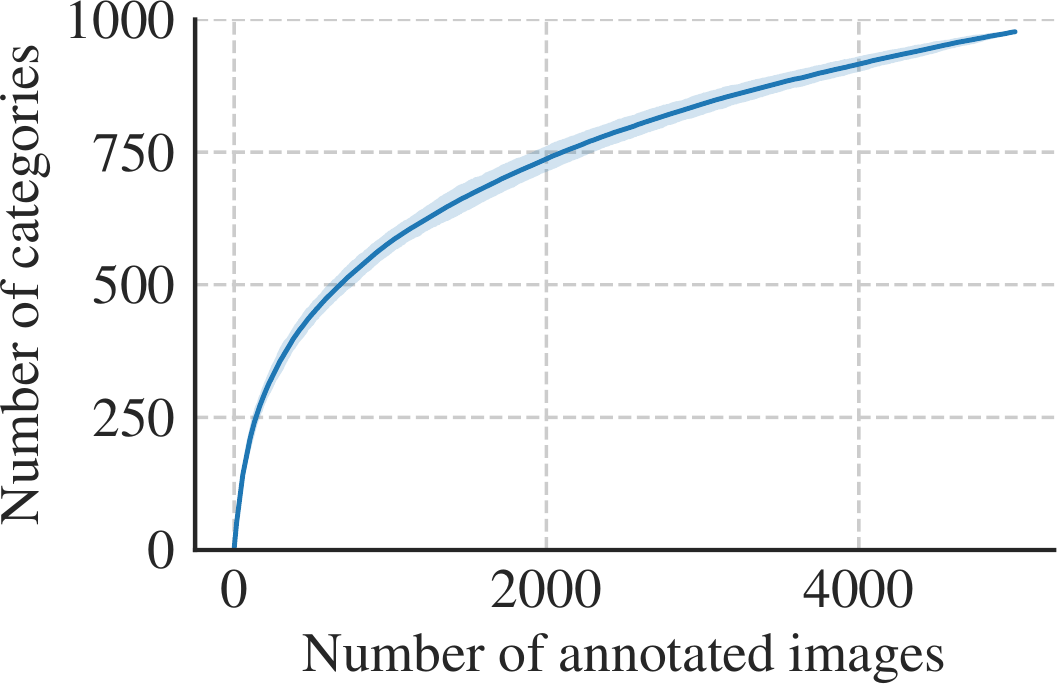}
\hspace{3mm}
\includegraphics[height=2.5cm]{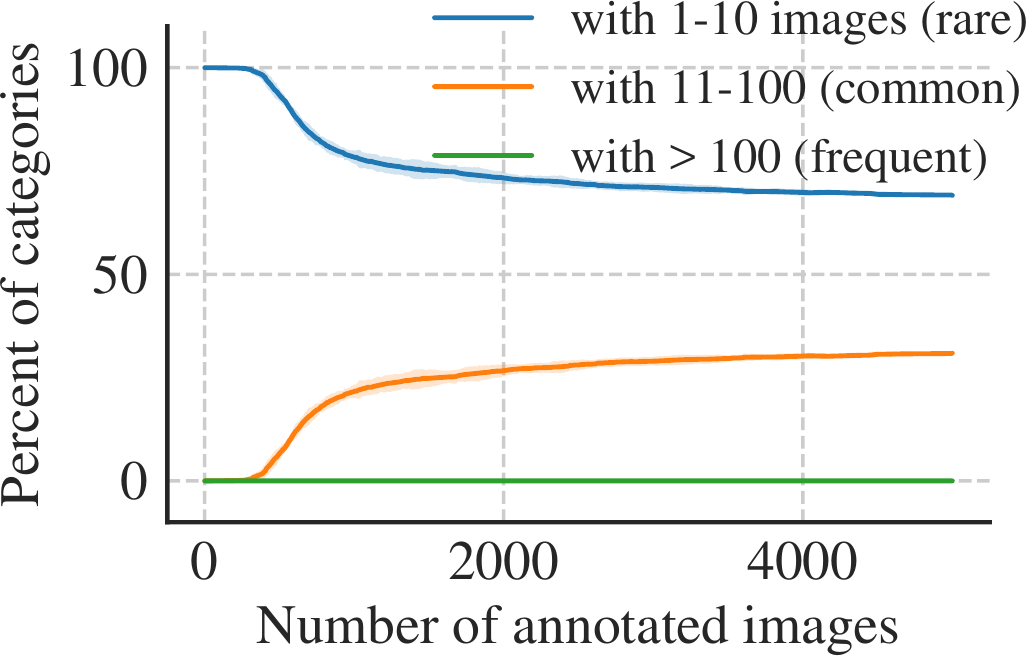}
\vspace{-.5mm}
\caption{(Left) As more images are annotated, new categories are discovered. (Right) Consequently, the percentage of low-shot categories (blue curve) remains large, decreasing slowly.}
\label{fig:analysis:cumulative_category_count}
\vspace{-5mm}
\end{figure}
%##################################################################################################

\paragraph{Low-Shot Category Statistics.} Fig.~\ref{fig:analysis:cumulative_category_count} (left) shows category growth as a function of image count (up to 977 categories in 5k images). Extrapolating the trajectory, our final dataset will include over 1k categories (upper bounded by the vocabulary size, 1723). Since the number of categories increases during data collection, the low-shot nature of \lvis is somewhat independent of the dataset scale, see Fig.~\ref{fig:analysis:cumulative_category_count} (right) where we bin categories based on how many images they appear in: \emph{rare} (1-10 images), \emph{common} (11-100), and \emph{frequent} ($>$100). These bins, as measured \wrt the training set, will be used to present disaggregated AP metrics.

%%%%%%%%%%%%%%%%%%%%%%%%%%%%%%%%%%%%%%%%%%%%%%%%%%%%%%%%%%%%%%%%%%%%%%%%%%%%%%%%%%%%%%%%%%%%%%%%%%%
\section{Conclusion}\label{sec:discussion}
We introduced \lvis, a new dataset designed to enable, for the first time, the rigorous study of instance segmentation algorithms that can recognize a large vocabulary of object categories ($>$1000) and must do so using methods that can cope with the open problem of low-shot learning. While \lvis emphasizes learning from few examples, the dataset is not small: it will span 164k images and label $\app$2 million object instances. Each object instance is segmented with a high-quality mask that surpasses the annotation quality of related datasets. We plan to establish \lvis as a benchmark challenge that we hope will lead to exciting new object detection, segmentation, and low-shot learning algorithms.

\appendix

%##################################################################################################
\begin{figure}[t]\centering
\includegraphics[height=2.5cm]{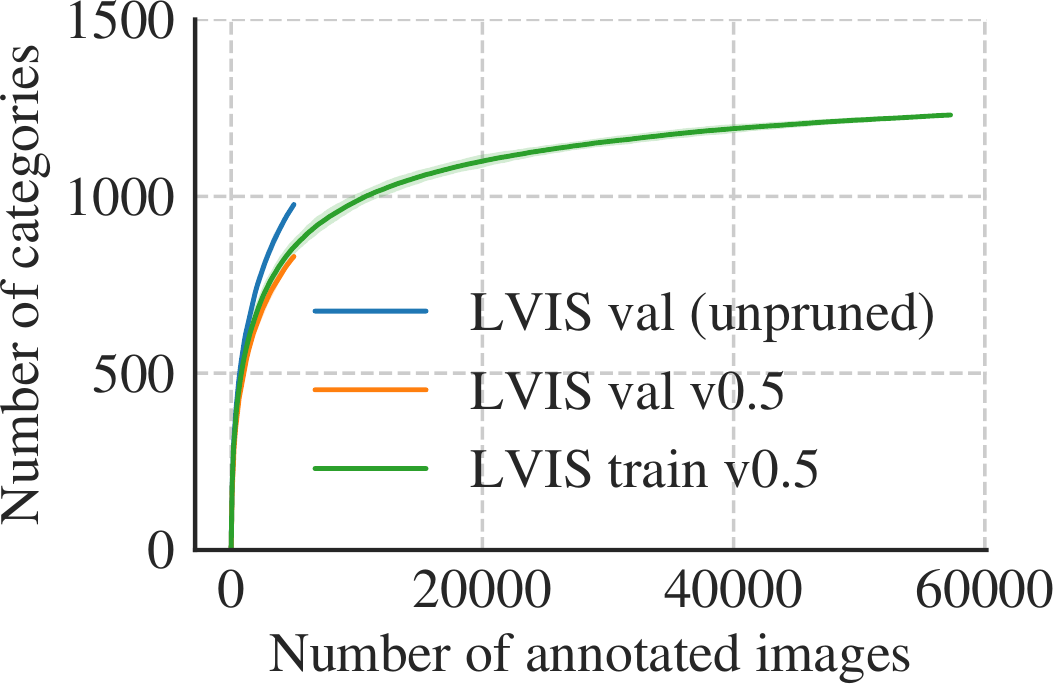}\hspace{3mm}
\includegraphics[height=2.5cm]{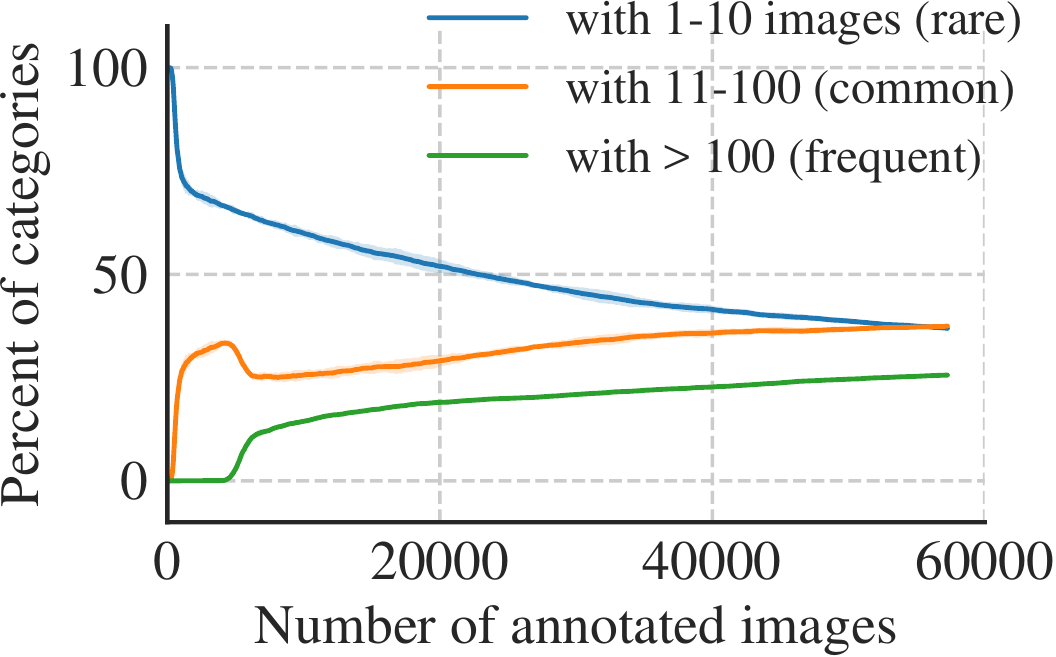}
\caption{Category growth (left) and frequency statistics (right) for \lvis v0.5. Best viewed digitally. Compare with Fig.~\ref{fig:analysis:cumulative_category_count}.}
\label{fig:analysis:cumulative_category_count_v0_5}
\end{figure}
%##################################################################################################

\section{\lvis Release v0.5}
\label{v0.5}

\lvis release v0.5 marks the halfway point in data collection. For this release, we have annotated an additional 77k images (57k \train, 20k \test) beyond the 5k \val images that we analyzed in the previous sections, for a total of 82k annotated images. Release v0.5 is publicly available at \url{https://www.lvisdataset.org} and will be used in the first \lvis Challenge to be held in conjunction with the COCO Workshop at ICCV 2019.

\paragraph{Collection Details.} We collected the v0.5 data in two 38.5k image batches using the process described in the main text. Each batch contained a proportional mix of \train and \test images. After collection was completed for the first batch, we manually checked all 1415 categories that were represented in the data collection and cast an include \vs exclude vote for each category based on its visual consistency. This process led to the removal of \app 18\% of categories and \app 10\% of labeled instances. After collecting the second batch, we repeated this process for 83 categories that were newly introduced. After we finish the full data collection for v1 (estimated early 2020), we will conduct another similar quality control pass on a subset of the categories.

\lvis \val v0.5 is the same as the set used for analysis in the main text, except that we removed any categories that: (1) were determined to be visually inconsistent in the quality control pass or (2) had zero instances in the training set. In this section, we refer to the annotations used for analysis in the main text as `\lvis val (unpruned)'.

\paragraph{Dataset Statistics.} After our quality control pass, the final category count for release v0.5 is 1230. The number of categories in the \val set decreased from 977 to 830, due to quality control, and it now has 51k segmented object instances. The \train v0.5 set has 694k segmented instances.

We next repeat some of the key analysis plots, this time showing the final \val and \train v0.5 sets compared to the original (unpruned) \val set from the main text. The \train and \test sets are collected using an identical process (the \train and \test images were originally sampled from the same image distribution and are mixed together in each annotation batch) and therefore the training data is statistically identical to that of the test data.

Fig.~\ref{fig:analysis:cumulative_category_count_v0_5} (left) illustrates the category growth rate on \train and \val before and after pruning. We expect only modest growth while collecting the second half of the dataset, perhaps expanding by \app 100 additional categories. Next, we extend Fig.~\ref{fig:analysis:cumulative_category_count} (right) from 5k images to 57k images using the \train v0.5 data, as shown in Fig.~\ref{fig:analysis:cumulative_category_count_v0_5} (right). Due to the slowing category growth, the percent of rare categories (those appearing in 1-10 training images) is decreasing, but remains a sizeable portion of the dataset. Roughly 75\% of categories appear in 100 training images or less, highlighting the challenging low-shot nature of the dataset.

\begin{figure}[t]\centering
\includegraphics[width=.75\linewidth]{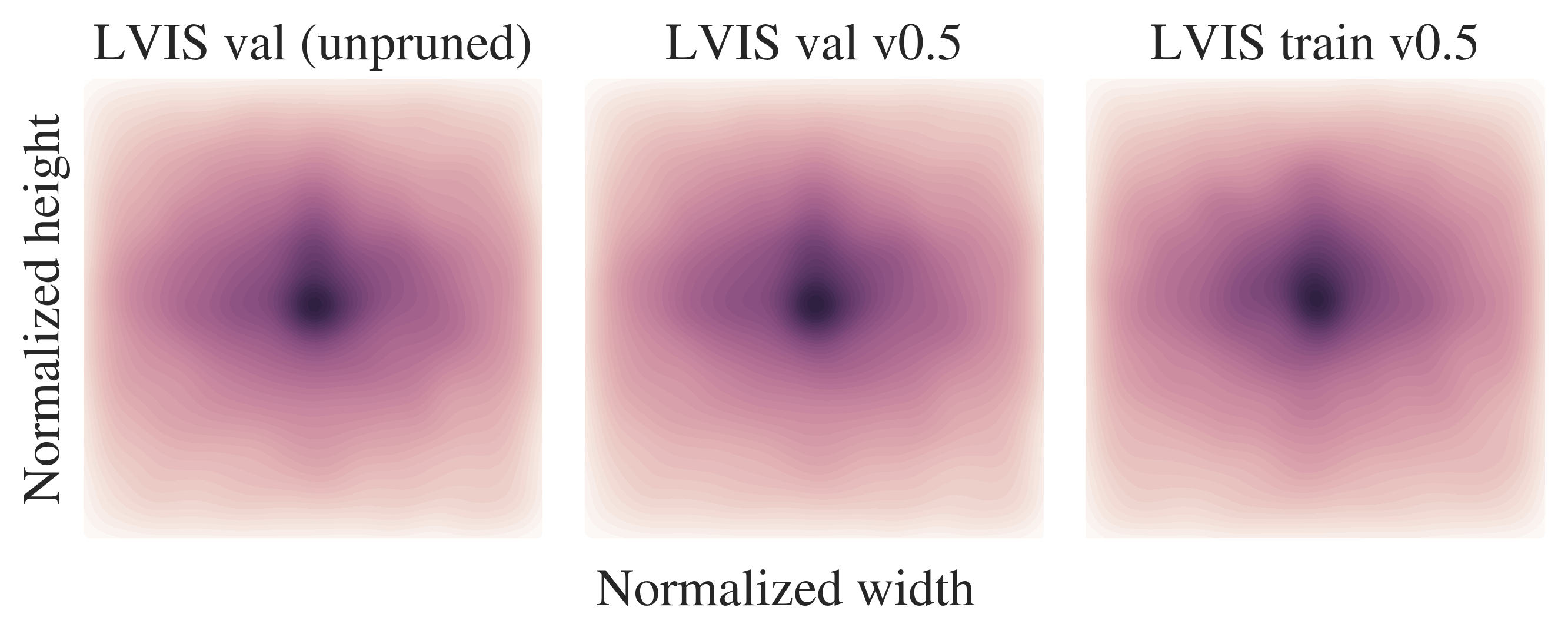}
\caption{Distribution of object centers in normalized image coordinates for \lvis \val, \lvis \val v0.5 (\ie after quality control), and \lvis \train v0.5. The distributions are nearly identical.}
\label{fig:analysis:center_distribution_v0_5}
\vspace{-3mm}
\end{figure}

Finally, we look at the spatial distribution of object centers in Fig.~\ref{fig:analysis:center_distribution_v0_5}. This visualization verifies that quality control did not lead to a meaningful bias in this statistic. The \train and \val sets exhibit visually similar distributions.

%##################################################################################################
%##################################################################################################

%##################################################################################################
\begin{table}[t]\centering
% subfloat a - out-of-the-box + inference tuning experiments
\subfloat[\label{tab:baselines:ootb}\textbf{Mask R-CNN baselines} (ResNet-50-FPN backbone). Top rows: adjusting two \emph{inference-time} hyper-parameters, the minimum score threshold and the number of detections per image, leads to a gain of 6.1 AP over the baseline using standard COCO hyper-parameters (row 1). The last two rows show the mean and standard deviation from five training runs.]{%
\tablestyle{4pt}{1.0}\begin{tabular}{@{}l@{\hspace{2mm}}l@{\hspace{1mm}}|cccc|c@{}}
  score thr & det/img & AP & AP$_\textrm{r}$ & AP$_\textrm{c}$ & AP$_\textrm{f}$ & AP$^\textrm{bb}$\\
  \shline
  0.050	& 100	& 14.8 & 0.8 & 10.9 & 25.3 & 14.8\\
  0.050	& 300	& 15.7 & 0.8 & 12.1 & 26.1 & 15.6\\
  0.001	& 300	& 20.8 & 3.3 & 20.7 & 27.9 & 20.3\\
  0.000	& 300	& 20.9 & 3.4 & 20.9 & 27.9 & 20.4\\
  \hline
  0.050	& 100 &	14.8\mypm{0.19} &	0.6\mypm{0.21} & 11.0\mypm{0.36} & 25.2\mypm{0.10}& 14.8\mypm{0.17}\\
  0.000	& 300	& \bf 21.0\mypm{0.17}	& \bf 3.2\mypm{0.35} & \bf 21.3\mypm{0.45} & \bf 27.7\mypm{0.12}& \bf 20.5\mypm{0.21}
\end{tabular}}\\
% subfloat b - data resampling
\subfloat[\label{tab:baselines:resampling}\textbf{Mask R-CNN with repeat factor sampling} (with best settings from Table~\ref{tab:baselines:ootb}). The frequency threshold $t$ controls the degree of resampling of rare categories ($t{=}0$ gives no resampling). Setting $t{>}0$ substantially improves AP$_\textrm{r}$ and $t{=}0.001$ gives best overall results. The last row presents class aware sampling (CAS), an alternate oversampling method~\cite{shen2016relay}.]{%
\tablestyle{12pt}{1.0}\begin{tabular}{@{}l|cccc@{}}
  $t$ & AP & AP$_\textrm{r}$ & AP$_\textrm{c}$ & AP$_\textrm{f}$ \\
  \shline
  0	& 21.0\mypm{0.17}	& 3.2\mypm{0.35} & 21.3\mypm{0.45} & \bf 27.7\mypm{0.12} \\
  0.0001 &21.2\mypm{0.14} &4.5\mypm{0.47} &21.5\mypm{0.37} & 27.6\mypm{0.14} \\
  0.0010 &\bf 23.2\mypm{0.21} &\bf 13.4\mypm{0.80} &\bf 23.2\mypm{0.32} &27.1\mypm{0.07} \\
  0.0100 &21.8\mypm{0.25} &9.8\mypm{1.27} &22.7\mypm{0.48} &25.6\mypm{0.13} \\
  0.1000 &21.3\mypm{0.24} &9.6\mypm{0.83} &21.7\mypm{0.32} &25.5\mypm{0.10} \\
  \hline
  CAS & 18.7\mypm{0.46} & 8.5\mypm{1.56} & 19.0\mypm{0.45} & 22.3\mypm{0.19} \\
\end{tabular}}\\
% subfloat c - enhancements
  \subfloat[\label{tab:baselines:enhancements}\textbf{Mask R-CNN enhancements}. We apply scale jitter data augmentation and upgrade the backbone to larger models~\cite{Xie2017}. This improves all AP metrics although AP$_\textrm{r}$ does not improve with the largest backbone.]{%
\tablestyle{4.8pt}{1.0}\begin{tabular}{@{}l|cccc@{}}
  enhancement & AP & AP$_\textrm{r}$ & AP$_\textrm{c}$ & AP$_\textrm{f}$ \\
  \shline
  Table~\ref{tab:baselines:resampling} best &23.2\mypm{0.21} &13.4\mypm{0.80} &23.2\mypm{0.32} &27.1\mypm{0.07} \\
  \; + scale jitter & 24.4\mypm{0.06} & 14.5\mypm{0.67} & 24.3\mypm{0.37} & 28.4\mypm{0.12} \\
  \; + ResNet-101 & 26.0\mypm{0.18} &\bf 15.8\mypm{0.95} & 26.1\mypm{0.21} & 29.8\mypm{0.22} \\
  \; + ResNeXt-101-32$\times$8d &\bf 27.1\mypm{0.43} &\bf 15.6\mypm{1.14} &\bf 27.5\mypm{0.77} &\bf 31.4\mypm{0.12} \\
\end{tabular}}\\[0.7em]
% main caption
\caption{\textbf{\lvis release v0.5 baselines.} Metrics: AP is mask AP; subscripts `r', `c', and `f' refer to rare, common, and frequent category subsets (defined in \S\ref{sec:analyis:eval}). Where applicable we repeat each experiment 5 times and report mean and standard deviation.}
\label{tab:baselines}
\vspace{-3mm}
\end{table}
%##################################################################################################

Based on this analysis and our qualitative judgement when performing per-category quality control, we conclude that our data collection process scales well beyond the initial 5k set analyzed in the main text.

\section{LVIS v0.5 Baselines}

To help researchers calibrate their results for the upcoming LVIS Challenge at ICCV 2019,\footnote{\url{https://www.lvisdataset.org/challenge}} we introduce simple baselines. In \S\ref{sec:baselines:ootb}, we test the performance of Mask R-CNN~\cite{He2017} out-of-the-box, and show the importance of adjusting two inference-time hyper-parameters. Next in \S\ref{sec:baselines:rep} we provide an improved (yet standard) baseline that resamples the training data in order to increase the frequency of rare categories. Finally in \S\ref{sec:baselines:enhance} we train larger models.

\subsection{Mask R-CNN Out-of-the-Box}\label{sec:baselines:ootb}
We first apply Mask R-CNN out-of-the-box on \lvis. Unless specified we use Mask R-CNN with a ResNet-50 backbone (pre-trained on ImageNet) with FPN~\cite{Lin2017}. Training is performed using Detectron2, which is implemented in PyTorch and will be open-sourced later this year. Our training formula is unmodified from COCO training.\footnote{We use SGD with 0.9 momentum and 16 images per minibatch; the training schedule is 60k/20k/10k updates at learning rates of 0.02/0.002/0.0002 respectively (this 90k update schedule is equivalent to \app 25 epochs over \train v0.5); we use a linear learning rate warmup~\cite{goyal2017accurate} over 1000 updates starting from a learning rate of 0.001; weight decay 0.0001 is applied; horizontal flipping is the only train-time data augmentation unless otherwise stated; training and inference images are resized to a shorter image edge of 800 pixels; no test-time augmentation is used.}

The results are low and in particular AP$_\textrm{r}$ (mask AP for rare categories) is 0.8\%---\emph{near zero}. In Table~\ref{tab:baselines:ootb} we demonstrate that adjusting two inference-time hyper-parameters on \lvis improves results. First, we increase the number of detections per images as \lvis allows up to 300 (\vs 100 for COCO). Second, due to class imbalance the max confidence scores reported for rare and common classes is typically low (compared to COCO), hence reducing the minimum score threshold from the default of 0.05 to 0.0 (\ie, no threshold) substantially improves AP$_\textrm{c}$. The combination of these changes increases AP$_\textrm{r}$ a modest amount, up to an average of 3.2\% Table~\ref{tab:baselines:ootb} (bottom row).

We additionally observe that on \lvis, \emph{mask AP is typically slightly higher than box AP} (denoted by AP$^\textrm{bb}$). This trend is the opposite for COCO, where AP is typically 3 to 4\% (absolute) lower than AP$^\textrm{bb}$. We hypothesize that the AP/AP$^\textrm{bb}$ trend on \lvis is due to high quality segmentation masks. We have found supporting evidence by programmatically degrading \lvis mask quality and observing a drop in AP with almost no change in AP$^\textrm{bb}$ (results not shown). %Indeed, degrading the masks to roughly COCO quality (as measured by vertices per polygon) resulted in a gap of \app2\% lower AP than AP$^\textrm{bb}$.

\subsection{Mask R-CNN with Data Resampling}\label{sec:baselines:rep}
Resampling training data is a common strategy for training models on class imbalanced datasets~\cite{shen2016relay,he2009learning,mahajan2018weaklysup,mikolov2013efficient}. We apply a method that was used to train large-scale hashtag prediction models in~\cite{mahajan2018weaklysup} (inspired by~\cite{mikolov2013efficient}). The method, which we refer to as \emph{repeat factor sampling}, increases the rate at which tail categories are observed by oversampling the images that contain them.

The method is implemented as follows. For each category $c$, let $f_c$ be the fraction of training images that contain at least one instance of $c$. Define the category-level repeat factor as $r_c = \max(1, \sqrt{t/f_c})$, where $t$ is a hyper-parameter. Since each image may contain multiple categories, we define an image-level repeat factor. Specifically, for each image $i$, we set $r_i = \max_{c\in i} r_c$, where $\{c \in i\}$ are the categories labeled in $i$. In each epoch, the SGD data sampler creates a random permutation of images in which each image is repeated according to its repeat factor $r_i$.

The one hyper-parameter of this method, $t$, is a threshold that intuitively controls the point at which oversampling kicks in. For categories with $f_c \le t$, there is no oversampling. For categories with $f_c > t$, the degree of oversampling follows a square-root inverse frequency heuristic: if we decrease the frequency of a category by a factor $\gamma< 1$, then its repeat factor will be multiplied by $\sqrt{1/\gamma}$. This heuristic has worked well in other settings, \eg~\cite{mikolov2013efficient}.

The results of repeat factor sampling for varying $t$ are shown in Table~\ref{tab:baselines:resampling}. Comparing with the baseline (equivalent to $t=0$), there is a large improvement in AP$_\textrm{r}$ from 3.2\% to 13.4\% at $t=0.001$. This threshold oversamples categories appearing in less than 0.1\% of images (829 of the 1230 categories). There is a slight penalty in lower AP$_\textrm{f}$ ($-$0.6\%), but overall AP improves ($+$2.2\%).

We also present results using \emph{class aware sampling} (CAS), a popular method on imbalanced classification datasets (\eg,~\cite{shen2016relay}). In CAS, the data sampler first selects a category and then an image containing that category. Consistent with repeat factor sampling and SGD best-practices~\cite{bottou2012stochastic}, we iterate over random permutations of categories and within each category random permutations of their images. CAS improves AP$_\textrm{r}$ over the baseline as expected (from 3.2\% to 8.5\%), however both AP$_\textrm{c}$ and AP$_\textrm{f}$ decrease leading to a worse overall result.

\subsection{Mask R-CNN Standard Enhancements}\label{sec:baselines:enhance}
Finally we consider two standard enhancements in addition to using repeat factor sampling with $t=0.001$: we apply scale jitter at training time (sampling image scale from \{640, 672, 704, 736, 768, 800\}) and upgrade to larger models. Both enhancement yield improvements as reported in Table~\ref{tab:baselines:enhancements} with a final validation AP of 27.1\%.

\section{\lvis \val to \test Results Transfer}

In this section we analyze how AP on \lvis v0.5 \val transfers to the \test set. First, in \S\ref{sec:appendix:category_bias} we show how the category distribution varies between the smaller \val and larger \test sets. Next, in \S\ref{sec:appendix:ap_bias} we demonstrate the surprising result that even with a fixed category set \emph{smaller evaluation sets can have a bias towards higher AP}. The impact is larger for rare categories, hence while it has a minimal effect on COCO, on \lvis it results in AP measured on the larger \test set to be lower than on the smaller \val set.

%##################################################################################################
\begin{figure}[t!]\centering
\includegraphics[height=3.2cm]{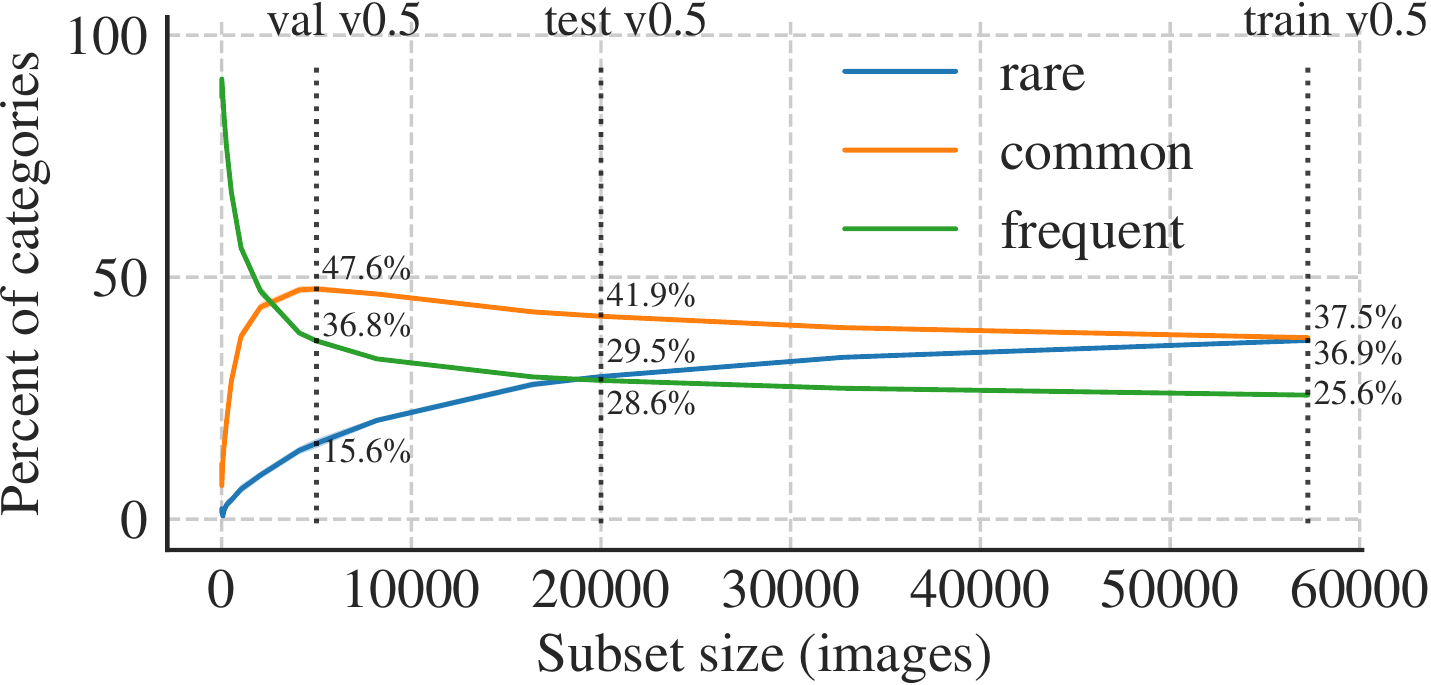}
  \caption{The distribution of rare, common, and frequent categories (defined \wrt \train v0.5) within random image subsets of a given size changes as a function of that size. The shaded region (imperceptible without zoom) illustrates one standard deviation around the mean over 10 draws of subsets for each size.}
\label{fig:analysis:r_c_f_dist}
\end{figure}
%##################################################################################################

\subsection{Category Frequency Distributions}\label{sec:appendix:category_bias}
An \emph{evaluation} set with a large proportion of categories that appear infrequently in the \emph{training} set (\ie categories with few training examples) will tend to be more difficult as learning from few examples is challenging. An important question, then, is what is the frequency distribution (\wrt the training set) of categories in a given evaluation set?

To investigate this question, we look at category frequency distributions in random image subsets of various sizes. For visualization, we quantize category frequency as described in \S\ref{sec:analyis:eval} into `rare', `common', and `frequent' groups based on how many images each category appears in the training set. In Fig.~\ref{fig:analysis:r_c_f_dist}, for each subset size, we plot the mean rare/common/frequent category distribution over random subsets of that size.

%##################################################################################################
\begin{figure*}[t!]\centering
  \subfloat[\label{fig:analysis:ap_bias_simulated}AP of simulated classifiers as a function of the evaluation set size and the fraction of positive examples $f_c$ (the number below each data point indicates the number of positives at that point, the shaded region indicates the standard error when averaged over 300 trials). The left plot shows the behavior of a random Gaussian classifier; the right shows a classifier that mimics the empirical score distribution of a trained classifier. While smaller $f_c$ leads to decreased AP, we also \emph{observe a consistent decrease in AP as the evaluation set size increases} (until convergence).]{\includegraphics[trim=-10 0 -10 0, clip, height=3.2cm]{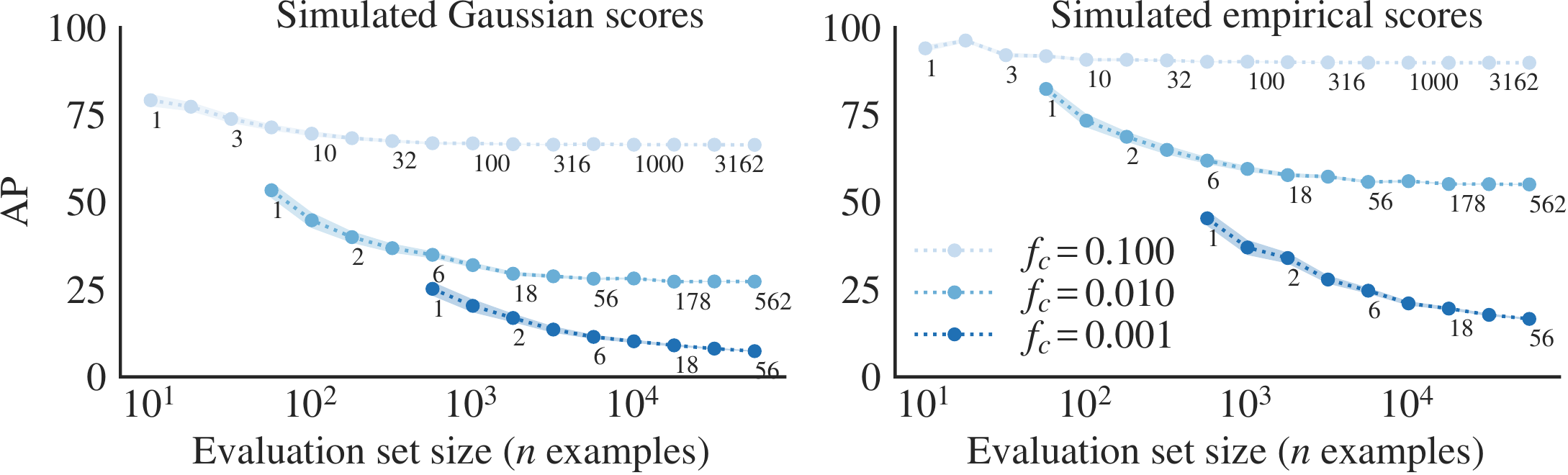}}
\hspace{0.5cm}
  \subfloat[\label{fig:analysis:ap_bias_coco}AP$^\textrm{bb}$@75 of a detector on three COCO categories when evaluated on random subsets of different sizes. Toasters are rare ($f_c=0.002$) while cats and dining tables appear more frequently. As in simulation, the AP can decreases with larger test set size, especially for rare categories.]{\includegraphics[trim=-20 0 -20 0, clip, height=3.2cm]{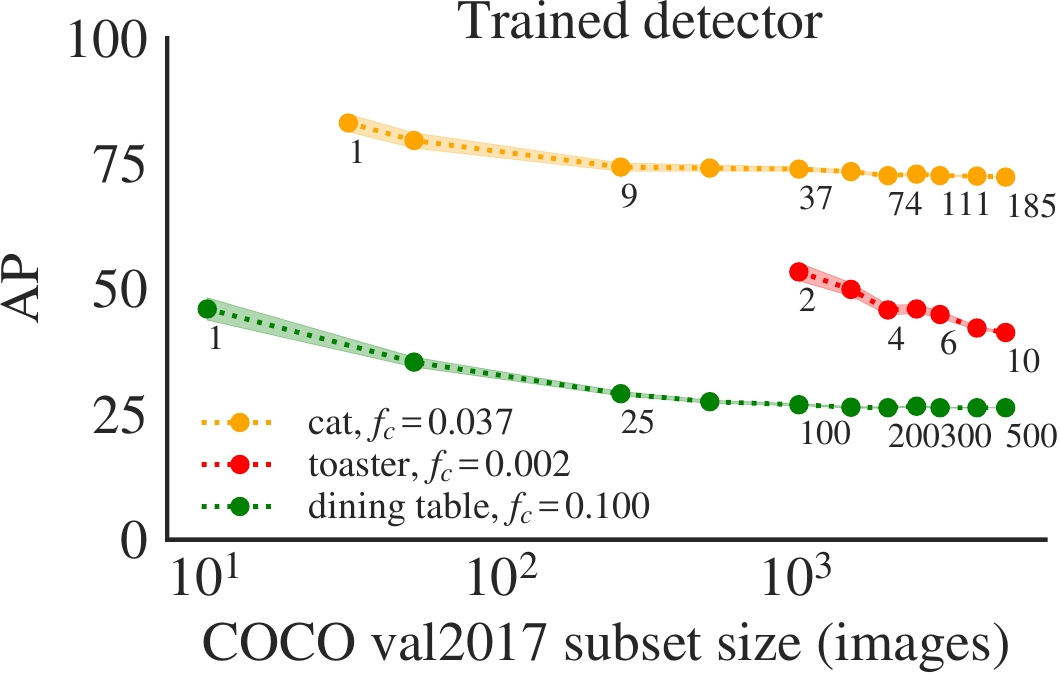}}
\vspace{2mm}
  \caption{\textbf{AP bias} as the size of the evaluation set is varied.}
\vspace{-3mm}
\end{figure*}
%##################################################################################################

%##################################################################################################
\begin{table*}[t!]\centering
% subfloat a - evaluation on random test subsets
\subfloat[\label{tab:val_test:subsets}Fixed Mask R-CNN model (Table~\ref{tab:baselines:resampling} best + scale jitter) evaluated on different size subsets of \test v0.5 (average over 30 random subsets). AP on the 5k subset is similar to AP on \val v0.5. As we increase subset size we observe a systematic decrease in all AP metrics consistent with the simulated and observed bias described in the main text.]{%
\tablestyle{10pt}{1.0}\begin{tabular}[t]{@{}l|cccc@{}}
  subset size & AP & AP$_\textrm{r}$ & AP$_\textrm{c}$ & AP$_\textrm{f}$ \\
  \shline
  5k & 24.8\mypm{0.51}& 11.5\mypm{1.71}& 25.5\mypm{0.86}& 30.1\mypm{0.28} \\
  10k & 22.1\mypm{0.31}& 10.5\mypm{0.66}& 22.9\mypm{0.56}& 29.2\mypm{0.20} \\
  15k & 20.8\mypm{0.23}& 10.0\mypm{0.54}& 21.7\mypm{0.37}& 28.9\mypm{0.11} \\
  \hline
  20k (full) & 18.4\phantom{\mypm{0.00}} & 8.8\phantom{\mypm{0.00}} & 18.7\phantom{\mypm{0.00}} &   27.2\phantom{\mypm{0.00}}
\end{tabular}}\hspace{6mm}
% subfloat b - transfer and ranking from val to test
\subfloat[\label{tab:val_test:transfer}We compare how AP transfers for three different models (Table~\ref{tab:baselines:enhancements}) from \val v0.5 to \test v0.5. All AP metrics decrease but the ranking of the models remains consistent across \val and \test.]{%
\tablestyle{8pt}{1.0}\begin{tabular}[t]{@{}l|c|cccc@{}}
  model & eval.\ set & AP & AP$_\textrm{r}$ & AP$_\textrm{c}$ & AP$_\textrm{f}$ \\
  \shline
  \multirow{2}{*}{ResNet-50} & \val & 24.4& 14.5& 24.3& 28.4 \\
  & \test & 18.4& 8.8& 18.7& 27.2 \\
  \hline
  \multirow{2}{*}{ResNet-101} & \val & 26.0& 15.8& 26.1& 29.8 \\
  & \test& 20.0& 9.4& 21.0& 28.7 \\
  \hline
  \multirow{2}{*}{ResNeXt-101-32$\times$8d} & \val& 27.1& 15.6& 27.5& 31.4\\
  & \test& 20.5& 9.8& 21.1& 30.0\\
\end{tabular}}\\[.5em]
% main caption
  \caption{\textbf{Results on \lvis \test v0.5} for different size subsets of \test and for three different baseline models.}
\label{tab:val_test}
\vspace{-3mm}
\end{table*}
%##################################################################################################

From this analysis, we can predict that the \val set (5k images) should contain \app 15\% rare categories, while the \test set (20k images) should have \app 29\% rare categories. Their actual values are 15.1\% and 28.2\%, validating our prediction. In general, larger evaluation sets contain a higher proportion of rare categories.
%\footnote{Note that for Fig.~\ref{fig:analysis:images_per_category_train_v0_5} the rare/common/frequent membership is defined \wrt the number of annotated images on the x-axis and therefore membership changes along the x-axis, while in Fig.~\ref{fig:analysis:r_c_f_dist} membership is statically determined by the full v0.5 \train set.}
Using the AP$_\textrm{r/c/f}$ achieved by R-101-FPN Mask R-CNN as an example, this distribution shift could result in a decrease in overall AP of \app2\% when moving from \val to \test. While optimizing for the LVIS Challenge, one may want to take this distribution shift into account; on the \test set rare categories will play a more important role than on the \val set.

\subsection{AP as a Function of Evaluation Set Size}\label{sec:appendix:ap_bias}
Suppose we have a small evaluation set (\eg, \val, 5k images) and a large evaluation set (\eg, \test, 20k images) that are both random samples from the same population. One might expect that while the categories present in the sets are different, the \emph{per-category AP} for a given category computed on evaluation sets of different sizes should be unbiased estimates of the true AP and only the variance of the estimate should change. Surprisingly, in general this intuition is not true and the estimate can be biased for smaller evaluation set sizes. We first observed this bias in Fig.~\ref{fig:analysis:pos}, in which we see that AP increases on average as the number of positive images per category ($|\posc|$) decreases. We now analyze this bias further in both simulated\footnote{The simulation code will be available on the \lvis website.} and real data.

Consider a simple \emph{binary classification} setting with $n$ test examples in total. Each example is positive for category $c$ with probability $f_c$. We will compute AP as a function of the evaluation set size $n$ and for various values of $f_c$. We consider two simulated classifiers, each of which is defined by class-conditional probability distributions over scores: $p_1(s) = p(s | y=1)$ and $p_0(s) = p(s | y=0)$, where $s$ is the classifier score and $y$ indicates the true label.

In the first simulation the classifier scores are drawn from $p_1(s) = \mathcal{N}(1,1)$ and $p_0(s) = \mathcal{N}(-1,1)$, \ie, the scores are Gaussian-distributed around $+1$ and $-1$, respectively. In the second simulation, we draw classifier scores from $p_1(s)$ and $p_0(s)$ obtained from a real classifier (the distributions, not shown, are highly non-Gaussian). Results for both simulations are shown in Fig.~\ref{fig:analysis:ap_bias_simulated}. Each curve shows the AP for a classifier as the evaluation set size (x-axis) varies. Different curves correspond to different category frequencies (0.001 to 0.1). Given a \emph{fixed} classifier, we observe that the curves are ordered top-to-bottom by higher-to-lower $f_c$. This ordering shows that for a fixed classifier, AP tends to be lower for rarer categories, which is an intuitive and well-known trend (see \cite{Hoiem2012} \S3.2). More surprising is the finding that within a curve, \emph{AP is consistently higher when the evaluation subset size is smaller.} This pattern exists at all frequency levels in both simulations.

Now moving from the simulated classification problem to real object detection data, we show AP$^\textrm{bb}$@75 of a trained detector for three categories evaluated on random COCO \texttt{val2017} subsets of various sizes in Fig.~\ref{fig:analysis:ap_bias_coco}. The toaster category is one of the two rarer categories in COCO while cats and and dining table appear more frequently. In each case we observe similar trends as in the earlier simulations.

Most categories in COCO are well-sampled like the cat and dining table categories and their AP has already converged on the 5k \texttt{val2017} set. Therefore overall AP does not vary much on COCO when comparing \texttt{val2017} to \texttt{test2017} results.

\lvis, unlike COCO, is not artificially balanced and therefore it contains a large number of rare categories. Therefore we expect to see a change in AP when moving from the small \val v0.5 set to the larger \test v0.5 set. We see exactly the predicted effects in Table~\ref{tab:val_test:subsets} where we report the results of a fixed Mask R-CNN model on various sized subsets of the \test set from 5k to 20k images. The results on the 5k subset size are in line with results on the 5k image \val set. As the number of evaluation images increases, we observe that AP systematically decreases.

Despite the bias between \val and \test (or more generally evaluation sets of different sizes), we expect the ranking of different models on the \val set and \test to remain constant under typical conditions as it is not obvious how one would exploit the bias. In Table~\ref{tab:val_test:transfer} we compare three models with distinctly different AP on the \val set to each other on the \test test. Indeed for at least these three models the ranking on \val transfers exactly to the ranking on \test.

\subsection{Comparing Models}
Finally, to gain a sense for the statistical differences between the three models in Table~\ref{tab:val_test:transfer}, we applied three standard hypothesis tests (paired t-test, random permutation test, and percentile bootstrap) to the mean of the per-category AP differences between pairs of models. As an example, for the ResNet-101 and ResNeXt-101 based models (\test AP of 20.0 \vs 20.5, respectively) the paired t-test and random permutation test return $p = 0.0490$ and $p = 0.0486$, respectively. The percentile bootstrapped produced a 95\% confidence interval of [0.002, 1.015], which excludes 0 (barely). These tests agree with each other (\ie, they reject, at a 5\% level, the null hypothesis that the mean difference in per-category AP values is zero) and provide some intuition for the statistical significance that might arise from a 0.5\% absolute difference in AP on \test.

\subsection{Summary}
In existing class-balanced detection datasets, researchers have grown accustomed to AP transferring nearly perfectly between small validation sets (\eg, 5k images) and larger test sets (20k images). In this section we demonstrated that when a dataset has a larger class imbalance there are at least two factors that cause AP estimated on smaller evaluation sets to be biased compared to larger evaluation sets. Empirically, this bias leads to higher AP on \val v0.5 than on \test v0.5. While a small validation set was unavoidable for \lvis v0.5, based on this analysis we may extend the validation set to include more images in release v1.

{\paragraph{Acknowledgements.}\small We would like to thank Ilija Radosavovic, Amanpreet Singh, Alexander Kirillov, Judy Hoffman, and Tsung-Yi Lin for their help during creation of LVIS\@. We thank the COCO Committee for granting us permission to annotate the COCO test set and Amanpreet Singh for help in creating the LVIS website.}

{\setstretch{.98}
\fontsize{8.5}{10}\selectfont
\bibliographystyle{ieee}
\bibliography{lvis}
}

\end{document}